\documentclass[
11pt,
a4paper,
oneside, 
headinclude,footinclude, 
BCOR5mm,
]{article}
\usepackage[utf8]{inputenc}
\usepackage{amsmath}
\usepackage{subcaption}
\usepackage{caption}
\usepackage{longtable}
\usepackage{amsthm}
\usepackage{color}
\usepackage{sectsty}
\usepackage{hyperref}
\hypersetup{
    colorlinks=true, % make the links colored
    linkcolor=blue, % color TOC links in blue
    urlcolor=red, % color URLs in red
    linktoc=all % 'all' will create links for everything in the TOC
}
\usepackage{amssymb}
\usepackage{pagecolor,lipsum}
\usepackage{enumerate}
\usepackage{pdfpages}
\usepackage{xcolor}
\usepackage{tikzsymbols}
\usepackage{blkarray}
\usepackage{enumitem}
\usepackage{graphicx}
\usepackage{setspace}
\usepackage{authblk}
\usepackage{soul}
\usepackage{enumitem}
\usepackage{soul}
\usepackage{MnSymbol}
\usepackage{mathtools}
\usepackage{braket}
\usepackage{mdframed}
\usepackage{amsmath}
\usepackage[thinc]{esdiff}
\usepackage{array}
\usepackage{booktabs}
\DeclarePairedDelimiter\ceil{\lceil}{\rceil}

\usepackage[linesnumbered,ruled,vlined,english]{algorithm2e}
\usepackage[noend]{algpseudocode}
\usepackage[linesnumbered,ruled,vlined,english]{algorithm2e}
\usepackage{geometry}
 \usepackage{tgtermes}

 \let\oldpara\paragraph
\renewcommand{\paragraph}[1]{\vspace{-0.5cm}\oldpara{#1}}

\newcommand{\mc}[1]{\mathcal{#1}}

\newcommand{\ep}[1]{\mathbb{E}}
\newcommand{\done}[1]{\textcolor{blue}{(\textbf{Done})}}

\newcommand{\pr}[1]{\mathbb{Pr}}

\newcommand{\innerthmname}{}% initialize

\theoremstyle{definition}

\newcommand{\z}[1]{\mathbb{Z}}
\newcommand{\nminfpe}[1]{\textsc{NMin-FPE}}
\newcommand{\nmaxfpe}[1]{\textsc{NMax-FPE}}
\newcommand{\nminfpr}[1]{\textsc{NMin-FPR}}
\newcommand{\nmaxfpr}[1]{\textsc{NMax-FPR}}

\newcommand{\zt}[1]{\mathbb{Z}_T}

\newcommand{\G}{\mathcal{G}}
\newcommand{\Gv}{\mathcal{V}}
\newcommand{\Ge}{\mathcal{E}}

\newcommand{\Q}{\mathcal{Q}_{\mathcal{M}, \mathcal{V}'}}

\newcommand{\Ndim}[1]{\text{Ndim}(#1)}
\newcommand{\C}[2]{\mathcal{C}_{(#1, {#2})}}
\newcommand{\sampcom}[1]{m_{#1}(\delta, \epsilon)}
\newcommand{\Ch}{h(\mathcal{C})}
\newcommand{\Cf}{h^*(\mathcal{C})}
\newcommand{\score}{\Gamma_i[\mc{C}, v]}
\newcommand{\pmac}{\textsf{PMAC}}
\newcommand{\pac}{\textsf{PAC}}
\newcommand{\hclass}{\mathcal{H}}
\newcommand{\rem}{\textbf{Remark.}}

\usepackage{float}

\newmdenv[backgroundcolor=gray!10, topline=false, bottomline=false, rightline=false, innerlinewidth=0.4pt, roundcorner=4pt,linecolor=black,innerleftmargin=6pt,
innerrightmargin=6pt,innertopmargin=3pt,innerbottommargin=6pt]{mybox2}

\newtheorem{theorem}{\textbf{Theorem}}[section]

\newtheorem{claim}{\textbf{Claim}}[theorem]
\newtheorem{lemma}[theorem]{\textbf{Lemma}}
\newtheorem{definition}[theorem]{\textbf{Definition}}

\newtheorem{observation}[theorem]{Observation}

\definecolor{mycolor}{rgb}{0.122, 0.435, 0.698}
\newmdenv[topline=false, bottomline=false, rightline=false, innerlinewidth=0.4pt, roundcorner=4pt,linecolor=black,innerleftmargin=6pt,
innerrightmargin=6pt,innertopmargin=1pt,innerbottommargin=6pt]{mybox}

\newmdenv[backgroundcolor=blue!5, topline=false, bottomline=false, rightline=false, leftline=false, innerlinewidth=0.4pt, roundcorner=4pt,innerleftmargin=10pt,
innerrightmargin=10pt,innertopmargin=10pt,innerbottommargin=10pt]{mybox3}

\definecolor{darkblue}{RGB}{0,0,76}
\title{\textbf{Efficient PAC Learnability of Dynamical Systems \\ Over Multilayer Networks}}

\author { \small
    Zirou Qiu,\textsuperscript{1,2}
    Abhijin Adiga, \textsuperscript{2}
    Madhav V. Marathe,\textsuperscript{1,2}
    S. S. Ravi,\textsuperscript{2,3}\\
    Daniel J. Rosenkrantz,\textsuperscript{2,3}
    Richard E. Stearns,\textsuperscript{2,3}
    Anil Vullikanti\textsuperscript{1,2}
}

\date{}

\begin{document}
\maketitle

% --------------------
%       Content      -
% --------------------
\vspace{-1cm}
\begin{abstract}
\textbf{Abstract.}
Networked dynamical systems are widely used as formal models of real-world cascading phenomena, such as the spread of diseases and information. Prior research has addressed the problem of learning the behavior of an unknown dynamical system when the underlying network has a single layer. In this work, we study the learnability of dynamical systems over \textit{multilayer} networks, which are more realistic and challenging. First, we present an efficient \pac{} learning algorithm with provable guarantees to show that the learner only requires a small number of training examples to infer an unknown system. We further provide a tight analysis of the Natarajan dimension which measures the model complexity. Asymptotically, our bound on the Nararajan dimension is tight for {\em almost all} multilayer graphs. The techniques and insights from our work provide the theoretical foundations for future investigations of learning problems for multilayer dynamical systems.

\smallskip
\noindent
\textbf{Conference version.} The paper is accepted at \texttt{\textbf{ICML-2024}}: \href{https://proceedings.mlr.press/v235/qiu24a.html}{\textbf{Link}} \footnote{Computer Science Dept., University of Virginia.; \textsuperscript{2} Biocomplexity Institute and Initiative, University of Virginia. \textsuperscript{3} Computer Science Dept., University at Albany – SUNY. Corresponding author: Zirou Qiu, \url{zq5au@virginia.edu}}. 
\end{abstract}

\section{Introduction}
Networked dynamical systems are mathematical frameworks for numerous cascade processes, including the spread of social behaviors, information, diseases, and biological phenomena~
\cite{battiston2020networks,ji2017mathematical,lum:jrsi-2014,sneddon2011efficient,schelling2006micromotives,LS-2004,Kauffman-etal-2003}. 
In general, such a system consists of an \textit{underlying graph} where vertices are entities (e.g., individuals, genes), and edges represent relationships between the entities. In modeling the contagion propagation, each vertex maintains a \textit{state} and a set of \textit{interaction functions} (i.e., behavior), which specify how the state evolves over time. Overall, the system dynamics proceeds in discrete time, with vertices updating their states using interaction functions. 

% Due to the extensive scale of real-world cascades, a complete specification of the underlying dynamical system is often not available. Towards this end, 
Inferring the unknown components of dynamical systems is an active research area~\cite{chen2022learning,dawkins2021diffusion,conitzer2020learning,adiga2019pac,lokhov2016reconstructing,narasimhan2015learnability}. One direction focuses on learning the unknown {\em interaction functions} of vertices. Interaction functions are crucial to the system dynamics; they specify the decision-making rules that entities employ to update their states. 
An illustrative example is the class of threshold interaction functions~\cite{granovetter1978threshold}, which are widely used to model the spread of social contagions~\cite{li2020online,watts2002simple}. Under this framework, each entity in the network has a decision threshold that represents the tipping point for a behavioral (i.e., state) shift. For instance, in rumor propagation, a person's belief changes when the number of neighbors believing in the rumor reaches a threshold~\cite{trpevski2010model}. Overall, the interaction functions define the mechanism of the cascade process, which also describes the system's global behavior~\cite{del2016spreading}. 

Existing methods for learning interaction functions only apply to the case when the underlying graph has a \textit{single-layer}. Nevertheless, such single-layer frameworks often are oversimplifications of reality, as real-world networks encompass diverse types of connections that are not adequately captured by single-layer graphs~\cite{kivela2014multilayer}.
To our knowledge, the learning problem for the more complex and realistic {\em multilayer} setting (that gives rise to multi-relational networks) has not received attention in the literature. In this work, we fill this gap through a formal study of \textbf{the learnability of dynamical systems over multilayer networks}.

\noindent
\textbf{The multilayer setting.} The graph in our target system consists of multiple layers with generally {\em different} set of edges in each layer. This is a classic setting for multilayer networks, and the capacity of such networks to model complex real-world phenomena has been widely recognized 
(e.g.,~\cite{hammoud2020multilayer,kivela2014multilayer,newman2018networks}). 
Notably, multilayer networks allow heterogeneous ties between vertices, with edges in each layer capturing a particular type of interaction (e.g., close friend, acquaintance)~\cite{newman2018networks,kivela2014multilayer,de2013mathematical}. Further, the multilayer framework enables the modeling of more realistic and complex cascades that involve {\em cross-layer 
interactions}~\cite{de2016physics,salehi2015spreading,boccaletti2014structure}. These cascades are characterized by increasingly intricate interaction functions of the system.

\noindent
\textbf{Problem description.} Consider a multilayer networked system where the interaction functions of vertices are \textit{unknown}. By inferring the missing functions, we aim to learn a system that captures the behavior of the true unknown system, with performance guarantees under the \textsf{P}robably \textsf{A}pproximately \textsf{C}orrect (\pac{}) model~\cite{valiant1984theory}. We learn from \textit{snapshots of the true system's dynamics}, a common scheme considered in related papers~(e.g., \cite{chen2021network,wilinski2021prediction,conitzer2020learning}).
To further measure the expressive power of the hypothesis class and characterize the sample complexity of learning, we examine the {\em Natarajan dimension}~\cite{natarajan1989learning} of the model, 
a well-known generalization of the VC dimension~\cite{vapnik2015uniform}.
Overall, we aim to address the following questions: $(i)$ {\em How to efficiently learn multilayer systems?} $(ii)$ {\em What is the complexity of our model for learning multilayer systems?}

\noindent
\textbf{Challenges.}~\label{para:challenge} 
The multilayer setting poses new challenges. First, the number of hypotheses grows exponentially in the number of layers, thus requiring a learner to search a much larger space. Further, a learner is tasked with extracting information from the complex \textit{cross-layer interactions}. For example, while the training data (snapshots of dynamics) indicates a vertex's state change, it \textit{does not} indicate which layer(s) triggered the change. Additionally, analyzing a learner's performance is challenging, as incorrect predictions could be caused by any combination of layers. The intertwined connections between vertices in the multilayer setting further complicate the analysis of model complexity. Collectively, these factors distinctly set apart our problem from its single-layer counterpart. 

\noindent
\textbf{Our contributions are as follows:}
\begin{itemize}[leftmargin=*,noitemsep,topsep=0pt]
    \item[-] \textbf{Efficient learning.} We show that a small training set is sufficient to efficiently \pac{} learn a multilayer system. Specifically, we obtain the following results. $(i)$ We develop an efficient \pac{} learning algorithm with provable guarantees: w.p. at least $1 - \delta$, the prediction error is at most $\epsilon$, for any $\epsilon, \delta  > 0$. 
    $(ii)$ For any fixed $\epsilon$ and $\delta$, the number of training examples used by our algorithm is only $O(\sigma k \log{}(\sigma k))$, where $k$ is the number of layers and $\sigma$ is the number of vertices with unknown interaction functions. Thus, when $\sigma$ is fixed, the size of an adequate training set does \textit{not} increase with the network size or density. This result also provides an upper bound on the sample complexity that is tighter than the general information-theoretic bound~\cite{haussler1988quantifying}.
    $(iii)$ We extend the proposed learner to the \textsf{P}robably \textsf{M}ostly \textsf{A}pproximately \textsf{C}orrect  setting~\cite{balcan2011learning} and prove that the amount of training data can be further reduced when the learner is allowed to make approximated predictions. $(iv)$ Using real-world and synthetic multilayer networks, we experimentally explore the relationship between the \pac{} algorithm's performance and system parameters under various scenarios.
    
\item[-] \textbf{Model complexity.} We provide a tight analysis of the Natarajan dimension (Ndim), which measures the expressive power of the learning model. $(i)$ We present a novel combinatorial structure and establish its {\em equivalence} to shatterable sets. 
$(ii)$ Based on this connection, we develop an (efficient) method for constructing shatterable sets and show that when restricting the system to an individual layer, Ndim is \textit{exactly} $\sigma$, the number of vertices with unknown interaction functions. This precise characterization could be of independent interest. $(iii)$ We then extend the argument to show that for a $k$-layer system, Ndim is between $\sigma$ and $k\sigma$ and present classes of instances where the bounds are tight. This result also provides a lower bound on the sample complexity. $(iv)$ Lastly, using a probabilistic argument, we show that our upper bound $k\sigma$ is asymptotically tight almost surely: for {\em almost all graphs}, Ndim is exactly $k\sigma$.
\end{itemize}

\paragraph{Related work.} For learning {\em single-layer} networks, many researchers have developed methods to address 
problems related to {\em cascade inference} (e.g., learning the diffusion functions at vertices, edge parameters, source vertices, and contagion states of vertices) by observing propagation dynamics. 
For instance, Chen et al.~\cite{chen2021network} study the problem of learning both the edge probability and source vertices under the celebrated independent cascade model. In another work by Conitzer et al.~\cite{conitzer2020learning}, they examine the setting where the opinions (states) of vertices are to be learned in stochastic cascades under the \pac{} scheme (\pmac{} to be more precise). Lokhov \cite{lokhov2016reconstructing} investigates the issue of parameter reconstruction for a special diffusion model, where they learn from real cascades. Other representative examples on learning single-layer systems include~\cite{ chen2022learning, conitzer2022learning, daneshmand2014estimating, dawkins2021diffusion,du2014influence, bailon:nsr-2011, he2016learning, hellerstein2007pac, kalimeris2018learning, narasimhan2015learnability,wilinski2021prediction}. Questions on learning the network topology from the cascades have also been studied; see, for example~\cite{huang2019statistical, pouget2015inferring,abrahao2013trace,du2012learning,myers2010convexity, gomez2010inferring, soundarajan2010recovering}. To our knowledge, the problem of learning the interaction functions of networked {\em multilayer} systems has not been examined.

\par The paper that is most closely related work is by Adiga et al.~\cite{adiga2019pac} in ICML 2019, where the \pac{} learnability of threshold interaction functions in {\em single-layer} networked systems
is studied. They present a consistent learner when there are only positive examples and show the hardness of learning when negative examples 
are also included. They also bound the sample complexity based on the VC dimension. As mentioned earlier, the multilayer setting introduces new \textit{challenges} that do not arise in the single-layer setting. For these reasons, we note that the results and techniques in~\cite{adiga2019pac} cannot be directly applied to our multilayer setting, which requires new techniques developed in our work.

\section{Preliminaries}\label{sec:pre}
Our setting aligns with the existing research on learning networked systems. For the readers' reference, we have included a list of settings used in several related papers in the Appendix (Section~\ref{sse:settings}).

\subsection{Multilayer Networked Dynamical Systems}
We now present the Multilayer Dynamical Systems as a formal model for cascades on multilayer networks.

\noindent
\textbf{Multilayer networks.}
All the graphs considered are undirected. For any integer $k \geq 1$,
let $[k]$ denote the set $\{1, \ldots, k\}$. A {\em multilayer network}~\cite{kivela2014multilayer} is a sequence of graphs $\mc{M} = \{\G{}_i\}_{i=1}^k$, $\G{}_i = (\Gv{}, \Ge{}_i)$, where $\Gv{}$ is a set of $n$ vertices {\em shared} by all graphs in $\mc{M}$, and $\Ge{}_i$ is the set of edges in $\G{}_i$, $i \in [k]$.  The edge sets of graphs in $\mc{M}$ are generally {\em different}.
Overall, one can view $\mc{M}$ as a $k$-layer network where $\G{}_i \in \mc{M}$ is the $i$th layer. 

\noindent
\textbf{Dynamical systems.} Dynamical systems on multilayer networks are generalizations of systems over single-layer networks.
A {\em Multilayer Synchronous Dynamical System} (\textbf{MSyDS}) over the Boolean domain $\mathbb{B} = \{0, 1\}$ is a triple $h^* = (\mc{M}, \mc{F}, \Psi)$: 
\begin{itemize}[leftmargin=*,noitemsep,topsep=0pt]
    \item $\mc{M} = \{\G{}_i\}_{i=1}^k$ is an underlying multilayer network with $k$ layers. 
    Each vertex has a state from $\mathbb{B}$.
    
    \item $\mc{F} = \{f_{i,v} : i \in [k], v \in \Gv{}\}$ is a collection of functions, with $f_{i,v}$ denoting vertex $v$'s \textbf{interaction function} on the $i$th layer $\G{}_i$. 
    %Each $f_{i,v}$ specifies the effect on $v$ from the state of $v$ and its neighbors on the $i$th layer.
    
    \item $\Psi = \{\psi_v : v \in \Gv{}\}$ is a collection of functions, 
    with $\psi_v$ denoting the \textbf{master function} of vertex $v$.
    % The role of these functions is explained below.
\end{itemize}

\noindent
The system dynamics proceeds in discrete time. Starting from an initial configuration of vertex states, at each step, vertices update states {\em synchronously} using interaction functions and master functions. 
Specifically, for any $t \geq 0$, the state of each vertex $v$ at time $t + 1$ is computed as follows:

\begin{itemize}[leftmargin=*,noitemsep,topsep=0pt]
    \item For each $\G{}_i \in \mc{M}$, the interaction function $f_{i,v} \in \mc{F}$ is evaluated; the inputs are the time-$t$ states of vertices in $v$'s closed neighborhood (i.e., $v$ and its neighbors) in $\G{}_i$; $f_{i,v}$ then outputs a value in $\mathbb{B}$. This gives a $k$-vector $\mathbf{W}_v$ for each $v$, where $\mathbf{W}_v(i)$ is the output of $f_{i,v}$, $i \in [k]$.
    
    \item Next, the master function $\psi_v$ is evaluated, with $\mathbf{W}_v$ as the input. The output of $\psi_v$, which is a value in $\mathbb{B}$, is the \textbf{next state} of $v$ (i.e., its state at time $t+1$).
\end{itemize}

\noindent
\textbf{Interaction functions.} We focus on {\em threshold} interaction functions, a classic framework to model the spread of social contagions~\cite{rosenkrantz2022efficiently,chen2021network,watts2002simple,granovetter1978threshold}. In particular, each $v \in \Gv{}$ has an integer threshold $\tau_i(v) \in [0, \text{deg}_i(v) + 2]$ for each layer $\G{}_i$, $i \in [k]$; $\text{deg}_i(v)$ is the degree of $v$ in $\G{}_i$. The interaction function $f_{i, v} \in \mc{F}$ outputs $1$ if the number of active (i.e., state-1) vertices in $v$'s closed neighborhood in $\G{}_i$ is \textbf{at least} $\tau_i(v)$; $f_{i, v}$ outputs $0$ otherwise.
If $f_{i,v}$ outputs $1$, we say that the \textbf{threshold condition} of $v$ is satisfied on $\G{}_i$.

\noindent
\textbf{Master functions.} The two classes of master functions proposed in the literature are \texttt{OR} and \texttt{AND}~\cite{pastor2015epidemic,lee2014threshold,brummitt2012multiplexity}. When function $\psi_v$ is \texttt{OR}, the next state of $v$ is $1$ iff \textbf{there exists} a layer $i \in [k]$ where the interaction function $f_{i, v}$ evaluates to $1$. 
In other words, $v$'s next state is 1 iff its threshold condition is satisfied in at least one layer. Analogously, for \texttt{AND} functions, the next state of $v$ is $1$ iff $f_{i, v}$ evaluates to $1$ in all the layers. 

\noindent
\textit{An illustrative example.} Consider the scenario of a rumor spreading on a 2-layer social network with two types of ties (e.g., “friends” and “co-workers”). A person’s belief in the rumor changes if the number of neighbors in any one (i.e., the \texttt{OR} master function) of the layers believing in the rumor reaches a certain decision threshold; however, the person’s thresholds for the two layers may be different due to the difference in the strengths of the social ties.

\begin{figure}[!h]
\small
  \centering
\includegraphics[width=0.45\textwidth]{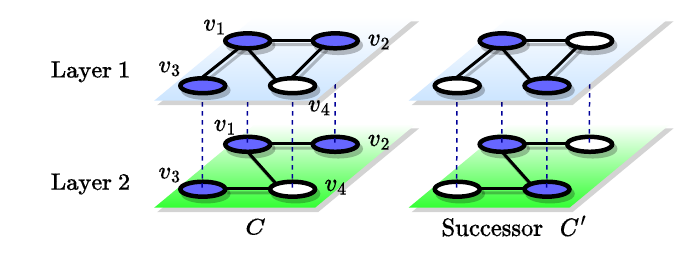}
    \caption{
        A 2-layer threshold system with \texttt{OR} master functions. Threshold values of vertices $v_1$ to $v_4$ in layer 1 are $(2, 3, 3, 2)$, and in layer 2 are $(3, 3, 2, 1)$. State-1 vertices are in blue. The configuration $\mc{C} = (1, 1, 1, 0)$, and its successor is $\mc{C}' = (1, 0, 0, 1)$.
    }
    \label{fig:example_OR}
\end{figure}

\noindent
\textbf{Configuration.} A \textbf{configuration} $\mc{C}$ is an $n$-bit binary vector that specifies the state of each vertex at a given time step. We use
$\mc{C}(v)$ to denote the state of vertex $v$ in $\mc{C}$. 
A configuration $\mc{C}'$ is the \textbf{successor} of $\mc{C}$ under a system $h^*$ if $\mc{C}'$ evolves from $\mc{C}$ in one time step; this is denoted by $\mc{C}' = h^*(\mc{C})$.
Overall, the evolution of system $h^*$ can be represented as a time-ordered sequence of configurations. An example of an evolution from $\mc{C}$ to $\mc{C}'$ is shown in Fig.~\ref{fig:example_OR}. We note that the goal of this figure was only to present a toy example of a multilayer system that makes it easier for readers to understand the formal definitions. As for examples of realistic large multilayer systems, we refer the reviewer to references such as~\cite{kivela2014multilayer} and~\cite{hammoud2020multilayer}.

\subsection{The Learning Setting}
There is a ground truth MSyDS $h^*$. The learner is only given partial information about $h^*$, where the interaction functions (on all layers) of a subset $\mc{V}' \subseteq \mc{V}$ of vertices are \textbf{unknown}.
Let $\sigma = |\mc{V}'|$. The {\em hypothesis class} $\hclass{}$ consists of $\Theta(n^{\sigma k})$ MSyDSs over all possible threshold values of vertices in $\mc{V}'$. The goal of a learner is to infer an MSyDS $h \in \hclass{}$ that is a good approximation of $h^*$ by inferring the unknown interaction functions. When $\mc{V}' = \mc{V}$, the thresholds of all vertices must be learned. 

\noindent
\textbf{Training.} We learn the target system $h^*$ from snapshots of its dynamics under the \pac{} framework. Formally, let $\mc{X} = \{0, 1\}^n$ be the set of all $n$-bit binary vectors. Let $\mc{T} = \{(\mc{C}_j, \mc{C}'_j)\}_{j = 1}^{q}$ be a {\em training set} of $q$ examples, which consists of the snapshots of system dynamics. Following the \pac{} setting, examples in $\mc{T}$ are {\em configuration pairs}, where $\mc{C}_j$ is drawn i.i.d. from an {\em unknown} distribution $\mc{D}$, and $\mc{C}_j' = h^*(\mc{C}_j)$ is the successor of $\mc{C}_j$ under $h^*$.  We use $\mc{T} \sim \mc{D}^q$ to denote such a training set. 

Intuitively, the training set $\mc{T}$ comprises snapshots of the freely-evolved system dynamics. The learner is \textbf{not} involved in generating $\mc{T}$. For instance, learning from a \textit{trajectory} can be seen as a special case of our \pac{} scheme. Specifically, when $\mc{T}$ consists of configurations on a trajectory $Y$, it suggests a special sampling distribution (unknown to the learner), where only configurations on $Y$ are sampled with positive probability. Importantly, our proposed learner works under {\em any} sampling distribution. 

% For a concrete example, consider a system that evolves freely. The training set consists of snapshots of the system's dynamics where the snapshots are sampled (by some entity) from an unknown distribution. Then, this sampled dynamics (i.e., the training set) is provided to the learner in a single batch. 
% The construction suggested by the reviewer, where each configuration is sampled and fed to the system, then the system outputs its successor, is just a special case of the sampling scheme considered in our work. 
% For example, the training set could also be constructed by sampling (the distribution is unknown to the learner) a pre-existing trajectory of the system evolution. This is a valid sampling distribution where all configurations on this trajectory are sampled with some probability, and any configuration not on this trajectory is sampled with probability zero.

\noindent
\textbf{Predictions.} Given a new sample $\mc{C} \sim \mc{D}$, a hypothesis $h \in \mc{H}$ makes a successful prediction if $h(\mc{C}) = h^*(\mc{C})$. The \textbf{true error} of a hypothesis $h$ is defined as $L_{(\mc{D}, h^*)}(h) := \Pr{}_{\mc{C} \sim \mc{D}} [h(\mc{C}) \neq h^*(\mc{C})]$. In the \pac{} model, when $\mc{T}$ is sufficiently large, the goal of a learner is to output an $h \in \mc{H}$ s.t. with probability at least $1 - \delta$ over $\mc{T} \sim \mc{D}^q$, it holds that $L_{(\mc{D}, h^*)}(h) \leq \epsilon$, for any given $\epsilon, \delta \in (0, 1)$. The minimum number of training examples needed by any \pac{} learner to learn $\mc{H}$ is called the {\em sample complexity} of $\mc{H}$.

\noindent
\textbf{Natarajan dimension.} Learning a hypothesis $h \in \hclass{}$ in our context can be cast as a {\em multiclass classification} problem, where $h$ maps a configuration $\mc{C}$ to one of the possible $2^n$ configurations (classes). To characterize the {\em model complexity} and the {\em expressive power} of the hypothesis class $\mc{H}$, we turn to the Natarajan dimension~\cite{natarajan1989learning}, which extends the VC dimension to multiclass settings. Formally, the \textbf{Natarajan dimension} of $\mc{H}$, denoted by $\Ndim{\mc{H}}$, is the maximum size of a {\em shatterable} set. A set $\mc{R} \subset \mc{X}$ is \textbf{shattered} by $\mc{H}$ if for every $\mc{C} \in \mc{R}$, there are two associated configurations, $\mc{C}^A, \mc{C}^B \in \mc{X}$, s.t. $(i)$ $\mc{C}^A \neq \mc{C}^B$, and $(ii)$ for every subset $\mc{R}' \subseteq \mc{R}$, there exists $h \in \mc{H}$ where $\forall \mc{C} \in \mc{R}', \; 
h(\mc{C}) = \mc{C}^A$ and $\forall \mc{C} \in \mc{R} \setminus \mc{R}', \; h(\mc{C}) = \mc{C}^B$.
% Further, $\Omega(\left(\Ndim{\mc{H}} + \log(1 / \delta)\right) / \epsilon)$ is a lower bound on the sample complexity of learning $\mc{H}$~\cite{Shwartz-David-2014}. Overall, the larger the value of $\Ndim{\hclass{}}$, the higher the expressive power of $\hclass{}$. 
%Zirou: Should we include the above?

\section{\pac{} Learnability of Multilayer Systems}\label{sec:pac}
In this section, we establish the efficient \pac{} learnability of the hypothesis class $\hclass{}$, defined in Section~\ref{sec:pre}. We first propose a learner that efficiently infers an unknown multilayer system. We then show that a training set of size $\ceil{{1}/{\epsilon} \cdot \sigma k \cdot \log{}({\sigma k}/{\delta})}$ is sufficient for the learner to achieve the $(\epsilon,\delta)$-\pac{} guarantee. Lastly, we prove that our algorithm can also handle the more general \pmac{} learning setting~\cite{balcan2011learning}, which permits learners to make approximate predictions. Due to space limits, \textbf{full proofs appear in the Appendix (Section~\ref{sse:sec-3})}. We present proofs for learning interaction functions under the \texttt{OR} master function. The results for \texttt{AND} master functions follow by duality (see Appendix, Section~\ref{sse:sec-3}). 

\subsection{An Efficient \pac{} Learner}
\label{sse:pac_alg}

For a configuration $\mc{C}$, a vertex $v \in \Gv{}$ and a layer $i \in [k]$, let $\Gamma_i[\mc{C}, v]$ be the number of $1$'s in the input to the interaction function $f_{i,v}$ under $\mc{C}$ (i.e., the number of state-1 vertices in $v$'s closed neighborhood in $\mc{G}_i$).
We call $\Gamma_i[\mc{C}, v]$ the \textbf{score} of $v$ in $\mc{G}_i$ under $\mc{C}$. Let $\tau^{h}_i(v)$ be the \textit{learned threshold} of $v$ for the $i$th layer in $h$, and let $\tau^{h^*}_i(v)$ be $v$'s \textit{true threshold} in the target system $h^*$.

\noindent
\textbf{\pac{} Learner.} Our algorithm learns a hypothesis $h \in \mc{H}$ by inferring the unknown thresholds in the target system $h^*$. Let $\mc{T} \sim \mc{D}^q$ be a given training set. For each vertex $v \in \mc{V}'$ with unknown thresholds on each layer $\mc{G}_i \in \mc{M}$, under \texttt{OR} master functions, we assign 

\begin{equation}
    \tau^{h}_i(v) = \max_{(\mc{C}, \mc{C}') \in \mc{T} :\, \mc{C}'(v) = 0} \{\Gamma_i[\mc{C}, v]\} + 1.
\end{equation}

If $\mc{C}'(v) = 1$~ for all $(\mc{C}, \mc{C}') \in \mc{T}$, we set $\tau^{h}_i(v) = 0$.

If the master function is \texttt{AND}, then we assign $\tau^{h}_i(v) = \min_{(\mc{C}, \mc{C}') \in \mc{T} :\, \mc{C}'(v) = 1} \{\Gamma_i[\mc{C}, v]\}$.
If $\mc{C}'(v) = 0$ for all $(\mc{C}, \mc{C}') \in \mc{T}$, we set $\tau^{h}_i(v) = \text{deg}_i(v) + 2$, where $\text{deg}_i(v)$ is the degree of $v$ in $\mc{G}_i$. 

Lastly, the algorithm returns the corresponding system $h \in \mc{H}$. One can easily verify that $h$ has zero empirical risk. Further, the running time is $O(n k \cdot |\mc{T}|)$.
Thus, the algorithm is an efficient consistent learner for $\mc{H}$.
Since $\hclass{}$ is finite, it follows that the class $\hclass{}$ is efficiently \pac{} learnable \cite{Shwartz-David-2014}.

\begin{mybox2}
\begin{theorem}
The class $\hclass{}$ is efficiently \pac{} learnable. 
\end{theorem}
\end{mybox2}

\subsection{The Sample Complexity}\label{sec:sample-complexity}

We now show that our algorithm requires a small training set to \pac{}-learn $\hclass{}$. To begin with, a well-known general result in~\cite{haussler1988quantifying} implies that the sample complexity $\sampcom{\hclass{}}$ of learning $\mc{H}$ is upper bounded by $(1/\epsilon) \log{}(|\hclass{}| / \delta)$, where $\epsilon, \delta \in (0, 1)$ are the two \pac{} parameters. In our case, one can derive that

\begin{equation}\label{eq:bound1}
     \sampcom{\hclass{}} ~\leq~ \frac{1}{\epsilon} \cdot \left(\sigma k \cdot \log{} (d_{\text{avg}}(\mc{V}')) + \log{} (\frac{1}{\delta})\right),
\end{equation}

where $\sigma = |\mc{V}'|$, and $d_{\text{avg}}(\mc{V}')$ is the average degree of the vertices of~$\mc{V}'$ in the network where layers are merged into a single layer with parallel edges removed. 

\noindent
\textbf{A new bound.} As one might expect, bound~(\ref{eq:bound1}) depends explicitly on the average degree since a denser network leads to a larger hypothesis class, requiring a larger training set. Nevertheless, we now establish an alternative bound on the training set size $|\mc{T}|$ for our algorithm. Notably, this bound {\em does not depend explicitly on any graph parameters} (e.g., $d_{avg}$) except for the number of layers $k$.

\noindent
\textbf{The key lemma.} 
For a configuration $\mc{C} \sim \mc{D}$, and a vertex $v$, let $B(\mc{C}, v)$ denote the event that ``$h^*(\mc{C})(v) = 0$'', i.e., in the true system $h^*$, $\mc{C}$ does not satisfy the threshold condition of $v$ on any layer. For a layer $i \in [k]$ and a system $h \in \hclass{}$, let $A(\mc{C}, i, v, h)$ be the ``bad'' event that $(1)$ ``the threshold condition of $v$ on the $i$th layer is satisfied under $h$'', and $(2)$ ``$B(\mc{C}, v)$ occurs''. Namely, $h$ makes a wrong prediction for the state of $v$ in $\G{}_i$. Formally, $A(\mc{C}, i, v, h)$ is defined as ``$\score{} \geq \tau^{h}_i(v)$ and $\forall j \in [k], \, \Gamma_j[\mc{C}, v] < \tau^{h^*}_j(v)$''.

We now bound the probability (over $\mc{T} \sim \mc{D}^q$) of our algorithm learning a ``bad'' $h \in \hclass{}$ s.t. $\Pr_{\mc{C} \sim \mc{D}}[A(\mc{C}, i, v, h)] \geq \alpha$, for a given $\alpha \in (0, 1)$.

\begin{mybox2}
\begin{lemma}\label{lemma:sample}
    For a $v \in \mc{V}'$ and an $i \in [k]$, suppose $\tau^{h^*}_i(v) \geq 1$. Let $h \in \hclass{}$ be a hypothesis learned from a training set $\mc{T}$ of size $q \geq 1$. For a given $\alpha \in (0, 1)$
    
    $(1)$ If all integers $\rho_i(v) \in [0, \tau^{h^*}_i(v))$ satisfy $\Pr_{\mc{C} \sim \mc{D}}[B(\mc{C}, v)~\mathrm{and}~\Gamma_i[\mc{C}, v] \geq \rho_i(v)] < \alpha$. Then $\Pr_{\mc{C} \sim \mc{D}}[A(\mc{C}, i, v, h)] < \alpha$. 

    $(2)$ If $(1)$ does not hold, that is, there is a $\rho_i(v) \in [0, \tau^{h^*}_i(v))$ such that $\Pr_{\mc{C} \sim \mc{D}}[B(\mc{C}, v)~ \mathrm{and}~\Gamma_i[\mc{C}, v] \geq \rho_i(v)] \geq \alpha$, then $\Pr_{\mc{C} \sim \mc{D}}[A(\mc{C}, i, v, h)] \geq \alpha$
    holds with probability \textbf{at most} $(1 - \alpha)^q$ over $\mc{T} \sim \mc{D}^q$.
\end{lemma}
\end{mybox2}

For an error rate $\alpha \in (0, 1)$, Lemma~\ref{lemma:sample} states that the probability (over $\mc{T} \sim \mc{D}^q$) of the algorithm learning a ``bad'' hypothesis $h$ where $\Pr_{\mc{C} \sim \mc{D}}[A(\mc{C}, i, v, h)] \geq \alpha$ is at most $(1 - \alpha)^q$. Using this lemma, we now present the result on the size of an adequate training set for our algorithm.

\begin{mybox2}
\begin{theorem}\label{thm:sample}
For any $\epsilon, \delta \in (0, 1)$, with a training set of size $q = \ceil{{1}/{\epsilon} \cdot \sigma k \cdot \log{}({\sigma k}/{\delta})}$, the proposed algorithm learns a hypothesis $h \in \hclass{}$ such that with probability at least $1 - \delta$ (over $\mc{T} \sim \mc{D}^q$), 
$
\Pr_{\mc{C} \sim \mc{D}}[h(\mc{C}) \neq h^*(\mc{C})] < \epsilon.
$
\end{theorem}
\end{mybox2}

\noindent
\textbf{Proof sketch.} 
We first prove that the learned hypothesis $h$ makes a mistake on a vertex $v$ if and only if $\Ch{}(v) = 1$ and $\Cf{}(v) = 0$, 
which corresponds to the event $A(\mc{C}, i, v, h)$ for some~$i \in [k]$. 
Then, using Lemma~\ref{lemma:sample}, we show that with probability at most $k \cdot (1 - \epsilon / (\sigma k))^q$ over $\mc{T} \sim \mc{D}^q$, the loss satisfies $\Pr{}_{\mc{C} \sim \mc{D}}[\Ch{}(v) \neq \Cf{}(v)] \geq {\epsilon}/{\sigma}$. It follows that with probability (over $\mc{T} \sim \mc{D}^q$) at most $\sigma k \cdot (1 - \epsilon/(\sigma k))^q$, we have $$\Pr{}_{\mc{C} \sim \mc{D}}[\Ch{} \neq \Cf{}] \geq \epsilon$$ 
Setting $$q = \ceil{\frac{1}{\epsilon} \cdot \sigma k \cdot \log{}(\frac{\sigma k}{\delta})}$$ one can verify that $\sigma k \cdot (1 - \epsilon/(\sigma k))^q \leq \delta$. Equivalently, with probability at least $1 - \delta$ over $\mc{T} \sim \mc{D}^q$, we have $\Pr{}_{\mc{C} \sim \mc{D}}[\Ch{} \neq \Cf{}] < \epsilon$. \qed

\noindent
\textbf{Implication on the sample complexity.} Theorem~\ref{thm:sample}
provides an upper bound on the sample complexity $\sampcom{\hclass{}}$ 
of learning $\hclass{}$. Specifically, we have
\begin{equation}
\sampcom{\hclass{}} \leq \ceil{\frac{1}{\epsilon} \cdot \sigma k \cdot \log{}(\frac{\sigma k}{\delta})}.
\end{equation}

\noindent
\rem{} With the proposed learner, Theorem~\ref{thm:sample} shows that an adequate number of examples to \pac{}-learn $\hclass{}$ does \textit{not} explicitly depend on the average degree or the size of the multilayer graph. Thus, when other parameters are fixed, the sample complexity of learning $\hclass{}$ does not grow as the graph increases in size or density 
(even though $\hclass{}$ itself grows exponentially), making the algorithm scalable to larger networks. Further, our bound in Theorem~\ref{thm:sample} is tighter than Ineq~\eqref{eq:bound1} in several regimes. For instance, for a fixed $\sigma$, the bound in Ineq~\eqref{eq:bound1} grows as $d_{avg}(\mc{V}')$ gets larger; on the other hand, our bound remains unchanged.

\subsection{Extension to the \pmac{} Model}\label{sec:pmac}
Our learner can operate in a more general \textsf{P}robably \textsf{M}ostly \textsf{A}pproximately \textsf{C}orrect (\pmac{}) framework~\cite{balcan2011learning}. In this setting, a learner aims to make predictions that are good {\em approximations} for the true values, allowing small errors in the predictions. The \pmac{} model is used in many contexts such as learning submodular functions~\cite{rosenfeld2018learning, balcan2011learning} and cascade inference~\cite{conitzer2020learning}.

\noindent
\textbf{Formulation.} In \pmac{} model, the learned hypothesis $h$ makes a successful prediction if $h(\mc{C})$ agrees with $h^*(\mc{C})$ on the states of more than $(1 - \beta)$ fraction of the vertices in $\Gv{}'$, for a given approximation factor $\beta \in (0, 1)$. Formally, let $W(h(\mc{C}), h^*(\mc{C}))$ be the number of vertices in $\mc{V}'$ whose states in $h(\mc{C})$ are {\em different} from those in $h^*(\mc{C})$. For given $\epsilon, \delta, \beta \in (0, 1)$, the goal is to learn a hypothesis $h \in \hclass{}$ such that with probability at least $1 - \delta$ over $\mc{T} \sim \mc{D}^q$, $\Pr{}_{\mc{C} \sim \mc{D}}[W(h(\mc{C}), h^*(\mc{C})) \geq \beta \sigma] ~\leq~ \epsilon$, where ``$W(h(\mc{C}), h^*(\mc{C})) \geq \beta \sigma$'' is the ``bad'' event that $h(\mc{C})$ does not approximate $h^*(\mc{C})$ to the desired factor.

We prove that under \pmac{} setting, the size of the training set for our algorithm (in Section~\ref{sse:pac_alg}) to learn an unknown system is significantly reduced.

% \par \textbf{Formulation.} In the \pmac{} model, for a $\mc{C} \sim \mc{D}$, the learned hypothesis $h$ makes a successful prediction if $h(\mc{C})$ agrees with $h^*(\mc{C})$ on the states of more than $(1 - \beta)$ fraction of the vertices in $\Gv{}'$, for a given factor $\beta \in (0, 1)$. Formally, let $W(h(\mc{C}), h^*(\mc{C}))$ be the number of vertices in $\mc{V}'$ whose states in $h(\mc{C})$ are {\em different} from those in $h^*(\mc{C})$. For given $\epsilon, \delta, \beta \in (0, 1)$, the goal is to learn a hypothesis $h \in \hclass{}$ such that with probability at least $1 - \delta$ over $\mc{T} \sim \mc{D}^q$,
% \begin{equation}
%     \Pr{}_{\mc{C} \sim \mc{D}}[W(h(\mc{C}), h^*(\mc{C})) \geq \beta \sigma] ~\leq~ \epsilon
% \end{equation}
% where ``$W(h(\mc{C}), h^*(\mc{C})) \geq \beta \sigma$'' is the ``bad'' event that $h(\mc{C})$ does not approximate $h^*(\mc{C})$ to the desired factor. 

% Our next theorem establishes the sample complexity to \pmac{} learn the hypothesis class $\hclass{}$.

\begin{mybox2}\label{thm:pmac-bound}
\begin{theorem}
For any given $\epsilon, \delta, \beta \in (0, 1)$, with a training set $\mc{T}$ of size $q = \ceil{{1}/{\epsilon} \cdot {1}/{\beta} \cdot k \cdot \log{} (\sigma k/\delta)}$, the proposed algorithm (Section~\ref{sse:pac_alg}) learns an $h \in \hclass{}$ such that with probability at least $1 - \delta$ over $\mc{T} \sim \mc{D}^q$, $h$ satisfies that $\Pr{}_{\mc{C} \sim \mc{D}}[W(h(\mc{C}), h^*(\mc{C})) \geq \beta \sigma] ~\leq~ \epsilon$.
\end{theorem}
\end{mybox2}

\noindent
\textbf{Proof sketch.} Let $A(\mc{C}, v, h)$ is the ``bad'' event where $\Ch{}(v) \neq \Cf{}(v)$ for a vertex $v \in \Gv{}'$ and $\mc{C} \sim \mc{D}$. Using Lemma~(\ref{lemma:sample}), where we set $\alpha = (\epsilon \beta) / k$, with probability (over $\mc{T} \sim \mc{D}^q$) at most $k \cdot (1 - (\epsilon \beta) / k)^q$, we have $\Pr{}_{\mc{C} \sim \mc{D}}[A(\mc{C}, v, h)] \geq (\epsilon \beta)$ for any vertex $v \in \Gv{}'$. Thus, with probability (over $\mc{T} \sim \mc{D}^q$) at most $\sigma k \cdot (1 - (\epsilon \beta) / k)^q$, there exists a vertex $v \in \mc{V}'$ such that $\Pr{}_{\mc{C} \sim \mc{D}}[A(\mc{C}, v, h)] \geq~ \epsilon \beta$. By the linearity of expectation, we have $$\mathbb{E}_{\mc{C} \sim \mc{D}}[W(h(\mc{C}), h^*(\mc{C}))] < \epsilon \beta \cdot \sigma$$ Then, using Markov Inequality, it follows that with probability (over $\mc{T} \sim \mc{D}^q$) at least $1 - \sigma k \cdot (1 - (\epsilon \beta) / k)^q$,  the learned $h \in \hclass{}$ satisfies $\Pr{}_{\mc{C} \sim \mc{D}}[W(h(\mc{C}), h^*(\mc{C})) \geq \beta \sigma] \leq \epsilon$. Lastly, Setting $$q = \ceil{{1}/{\epsilon} \cdot {1}/{\beta} \cdot k \cdot \log{} (\sigma k/\delta)}$$ we have $1 - \sigma k \cdot (1 - (\epsilon \beta) / k)^q ~\geq~ 1 - \delta$. 
\qed

\noindent
\rem{} Compared to the bound on the \pac{} sample complexity in Theorem~\ref{thm:sample}, the size of an adequate training set under the \pmac{} model is reduced by a multiplicative factor of $\sigma \beta$. For instance, when $\beta$ is a constant (i.e., to obtain a constant-factor approximation), the training set size is decreased by a linear (w.r.t. $\sigma$) factor. This demonstrates a trade-off between the quality of the prediction and the size of the training data: when our learner is allowed to make approximate predictions, it requires a much lower number of examples to learn an appropriate hypothesis.

\section{Tight Analysis of Model Complexity}\label{sec:nata}
The Natarajan dimension (Ndim) measures the expressiveness of a hypothesis class and characterizes the complexity of learning~\cite{natarajan1989learning}. The higher the value of Ndim, the greater the expressive power of the learning model. Further, Ndim informs us about the requisite sample size for learning good hypotheses. In this section, we examine Ndim of the hypothesis class $\hclass{}$ for multilayer systems, denoted by $\Ndim{\hclass{}}$, and develop the following results. \textbf{See Appendix (Section~\ref{sse:sec-4}) for full proofs.} 

\begin{itemize}[leftmargin=*,noitemsep,topsep=0pt]
\item[1.] We present an efficient method for constructing shatterable sets. Using this method, we establish that $\Ndim{\hclass{}}$ is \textit{exactly} $\sigma$ when the underlying network has only one layer. Previously, Adiga~et~al.~\cite{adiga2019pac} showed that for the single-layer case where
$\sigma=n$, the VC dimension of $\hclass$ is at least~$n/4$. Our precise characterization of $\Ndim{\hclass{}}$ provides both an improvement over their bound~\cite{adiga2019pac} and an extension to arbitrary~$\sigma$. 

\item[2.] We show that for multilayer networks, $\Ndim{\hclass}$ is bounded between $\sigma$ and $k \sigma$. This proof uses an extended version of our technique for the single-layer case. Further, we present classes of instances where the bounds are tight. Our results also show that the best possible lower bound of sample complexity that one can obtain using this approach is   $\Omega((\sigma + \log{}(1/\delta)) / \epsilon)$. 

\item[3.] We further tighten our analysis by proving that asymptotically, for \textbf{almost all} graphs (i.e., almost surely) with $n$ vertices and $k \geq 2$ layers, $\Ndim{\hclass}$ of the corresponding hypothesis class $\hclass{}$ is {\em exactly} $\sigma k$.
\end{itemize}

%%% Shortened version of previous text.
%%% Previous text.
%%Lastly, we further tighten our analysis by showing that 
%%in the space of all possible multilayer graphs with $n$ vertices %%and $k$ layers, the corresponding hypothesis class for at least %%$1 - O((kn)^2 \cdot (3/4))^n)$ proportion (tends to $1$ as $n 
%%\rightarrow \infty$) of them have $\text{Ndim} = \sigma k$. That %%is, asymptotically, for \textit{almost all} graphs, %%$\Ndim{\hclass}$ of the corresponding hypothesis class %%$\hclass{}$ is exactly $\sigma k$.
\subsection{An Exact Characterization for a Single Layer}
Let $h^*$ be the true system with a single-layer network. We present a combinatorial characterization of shatterable sets, which allows us to obtain an
exact value for $\Ndim{\hclass{}}$. We begin with some definitions.

\begin{definition}[\textbf{Landmark Vertices}]
Suppose the underlying network has a single layer. Given a 
set $\mc{R} \subseteq \mc{X}$, a vertex $v \in \mc{V}'$ is 
a \textbf{\textit{landmark}} vertex for a configuration 
$\mc{C} \in \mc{R}$ if the score $\Gamma[\mc{C}, v] \neq \Gamma[\Hat{\mc{C}}, v]$ for all $\Hat{\mc{C}} \in \mc{R} \setminus \{\mc{C}\}$.
\end{definition}

The landmark vertices play a key role in $\mc{R}$ being shatterable.
Given $\mc{R} \subseteq \mc{X}$, let $\mc{W}(\mc{R}) \subseteq \mc{V}'$ be the (possibly empty) set of vertices that are landmark vertices for at least one configuration in $\mc{R}$.

\begin{definition}[\textbf{Canonical Set}]\label{def:can-single} A set $\mc{R} \subseteq \mc{X}$ is \textbf{\textit{canonical}} w.r.t. $\hclass{}$ if there is an injective mapping from $\mc{R}$ to $\mc{W}(\mc{R})$ s.t. each $\mc{C} \in \mc{R}$ maps to a landmark vertex of $\mc{C}$.
\end{definition}

By the definition of shattering (see Section~\ref{sec:pre}), each $\mc{C}$ in a shatterable set $\mc{R}$ is associated with two configurations, denoted by $\mc{C}^A$ and $\mc{C}^B$, where $\mc{C}^A \neq \mc{C}^B$. 

\begin{definition}[\textbf{Contested Vertices}]\label{def:contested}
We call a vertex $v$ \textbf{contested} for a $\mc{C} \in \mc{R}$ if $\mc{C}^A(v) \neq \mc{C}^B(v)$.
\end{definition}

By linking landmark vertices to contested vertices, 
our next lemma shows that for a single-layer system, the property of a set being canonical is \textit{equivalent} to being shatterable.

\begin{mybox2}
\begin{lemma}\label{lemma:canonical}
When the underlying network has a single layer, a set $\mc{R} \subseteq \mc{X}$ can be shattered by $\hclass{}$ \textbf{if and only if} $\mc{R}$ is canonical w.r.t. $\hclass{}$.
\end{lemma}
\end{mybox2}

\noindent
\textbf{Proof sketch.} $(\Rightarrow)$ Suppose $\hclass$ shatters $\mc{R}$. We want to show that $\mc{R}$ is canonical. We first establish the following claims $(1)$ {\em All contested vertices are in $\mc{V}'$}. $(2)$ {\em No two configurations in $\mc{R}$ have a common contested vertex. Further, a \textbf{contested} vertex for a configuration $\mc{C} \in \mc{R}$ is also a \textbf{landmark} vertex for $\mc{C}$.} From the above claims, it follows that there exists an injective mapping from $\mc{R}$ to $\mc{W}(\mc{R})$; i.e., $\mc{R}$ is canonical. $(\Leftarrow)$ Suppose that $\mc{R} \subseteq \mc{X}$ is canonical w.r.t $\hclass{}$. To show that $\hclass{}$ shatters $\mc{R}$, we present a method to construct the associated configurations, $\mc{C}^A$ and $\mc{C}^B$, of each $\mc{C} \in \mc{R}$ by specifying the states of vertices. We then show that the shattering conditions are satisfied under our selected $\mc{C}^A$ and $\mc{C}^B$. \qed

\noindent
\rem{} By definition, the size of a canonical set is at most $|\Gv{}'| = \sigma$. From Lemma~\ref{lemma:canonical}, it follows that $\Ndim{\hclass{}}$, which is the maximum size of a shatterable set, is \textbf{at most} $\sigma$ when the underlying network has a single layer.

Next, we present an efficient method in Theorem~\ref{thm:exactly-sigma} for constructing a canonical set of size $\sigma$ based on depth-first search (see proof in the Appendix). 
Consequently, for \textit{any} underlying single-layer network, \textbf{there exists a shatterable set of size exactly} $\sigma$. Formally:
%Collectively, we obtain the exact result that for single-layer systems, $\Ndim{\hclass{}} = \sigma$.

\begin{mybox2}
\begin{theorem}\label{thm:exactly-sigma}
When the underlying network has a single layer, a shatterable set of size $\sigma$ can be (efficiently) constructed. Thus, $\Ndim{\hclass{}} = \sigma$.
\end{theorem}
\end{mybox2}

\noindent
\textbf{Proof sketch.} To construct a canonical set $\mc{R} \subset \mc{X}$ with $\sigma$ configurations, $\mc{R}$ is initially empty. Let $\mc{G}' = \mc{G}[\mc{V}']$ be the subgraph induced on $\mc{V}'$. Starting from any vertex $v_1 \in \mc{V}'$, the algorithm carries out a \textit{depth-first} traversal on $\G{}'$, while maintaining a stack $\mc{K}$ of vertices. Let $v_i$, $i \in [\sigma]$, denote the $i$th discovered vertex, $1 \leq i \leq \sigma$. When $v_i$ is discovered, the configuration $\mc{C}_{v_i}$ added to $\mc{R}$ is constructed as follows: $\mc{C}_{v_i}(v) = 1$ if $v \in \mc{K}$ and $\mc{C}_{v_i}(v) = 0$ otherwise. The algorithm terminates when all vertices in $\Gv{}'$ are visited. Clearly $|\mc{R}| = \sigma$. We then show that $v_i$ is a landmark vertex of $\mc{C}_{v_i}$, thereby establishing that $\mc{R}$ is canonical. By Lemma~\ref{lemma:canonical}, $\mc{R}$ is also shatterable.
\qed

\noindent
\rem{} Theorem~\ref{thm:exactly-sigma} is interesting since for many problems~\cite{vapnik1994measuring, bartlett2019nearly}, including those on learning networked systems~\cite{adiga2019pac}, known results on such dimensions provide only bounds rather than exact values. The graph-theoretic machinery we have introduced (e.g., canonical set) to aid our analysis may be of independent interest in studying other learning problems for dynamical systems.
%%% Dangerous; this machinery seems specific to threshold functions.
%Further, our developed machinery (e.g., canonical set) are of %independent interest, which may be used in other learning problems %for dynamical systems.

\subsection{Bounds on Ndim for Multilayer Systems}

Using the results developed in the previous section, we now derive bounds on $\Ndim{\hclass{}}$ for systems with $k \geq 2$ layers. We first prove that $\Ndim{\hclass{}}$ $\leq$ $k \sigma$ by showing that each $v \in \mc{V}'$ is \textit{contested} for at most $k$ configurations in any shatterable set. We then show that $\Ndim{\hclass{}} \geq \sigma$ by establishing that any shatterable set obtained by restricting the system to any single layer in $\mc{M}$ is also shatterable over the multilayer network $\mc{M}$. We first state the upper bound:

\begin{mybox2}
\begin{lemma}\label{lemma:at-most-k-sigma}
If the network has $k \geq 2$ layers, then the size of any shatterable set $\mc{R}$ is at most $k \sigma$.
\end{lemma}
\end{mybox2}

% \textbf{Proof sketch.}
% For a $v \in \mc{V}'$, let $\mc{R}_v \subseteq \mc{R}$ be the subset of configurations with $v$ being (one of) their contested vertices. W.l.o.g., suppose $\mc{R}_v \neq \emptyset$. We show the following: 
% \begin{claim}
%     For each $\mc{C} \in \mc{R}_v$, $\exists \; i \in [k]$ such that $\Gamma_i(\mc{C}, v) > \Gamma_i(\hat{\mc{C}}, v), \; \forall \; \Hat{\mc{C}} \in \mc{R}_v \setminus \{\mc{C}\}$.
% \end{claim}
% \midvs{}
% The claim implies that for each $\mc{C} \in \mc{R}_v$, there exists a layer $i$ where $v$'s score under $\mc{C}$ is strictly larger than $v$'s score under any other configurations in $\mc{R}_v$. Thus, $|\mc{R}_v| \leq k$, and $|\mc{R}| \leq k \sigma$. \qed

%%%% The following remark is not needed.
%%%% It was commented out by Ravi.
%\rem{} Despite the somewhat crude analysis to arrive at the %above bound, we latter show that this bound is %\textit{tight} for almost all the hypothesis classes for %threshold multilayer networked systems. 

\noindent
To establish a lower bound of $\sigma$ on the size of a shatterable set, we prove the following lemma.

\begin{mybox2}
\begin{lemma}\label{lemma:at-least-sigma}
Suppose $h^*$ is an MSyDS whose underlying network 
has $k \geq 2$ layers. Let $\Hat{h}^*$ be a single-layer system obtained from $h^*$ by using the network in any layer $i \in [k]$. If a set $\mc{R}$ is shatterable by the hypothesis
class of $\Hat{h}^*$, then it is also shatterable by the hypothesis class of $h^*$.
\end{lemma}
\end{mybox2}

By Theorem~\ref{thm:exactly-sigma}, when the underlying network has a single layer, there exists a shatterable set of size $\sigma$. Thus, Lemma~\ref{lemma:at-least-sigma} implies that there also exists a shatterable set of size $\sigma$ for the multilayer setting. Overall, we obtain the following bounds on $\Ndim{\hclass{}}$: 

\begin{mybox2}
\begin{theorem}\label{thm:bound-nata}
Suppose the underlying network has $k \geq 2$ layers. Then $\sigma \leq \Ndim{\hclass{}} \leq k \sigma$.
% \aacomment{Same as the previous theorem.}
\end{theorem}
\end{mybox2}

\noindent
\rem{} 
There are classes of multilayer systems where the bounds are tight. To match the lower bound, consider the class of $k$-layer networks $\mc{M} = \{\G{}_1, ..., \G{}_k \}$ where for each vertex $v$, the following condition holds: {\em the vertex $v$'s neighbors in $\G{}_{i+1}$ form a superset of its neighbors in $\G{}_i$, with exactly one extra neighbor in $\G{}_{i+1}$, $i = 1, ..., k-1$.} It can be verified that in such a system, given any shatterable set $R$, each vertex with unknown interaction functions can be contested for at most one configuration in this set. Therefore, $|R|$ is at most $\sigma$ and Ndim $= \sigma$. On the other hand, in Section~\ref{sec:asymptotic}, we show that the upper bound $k \sigma$ is tight for {\em almost all} threshold multilayer systems. 

\noindent
\textbf{Bounds on the sample complexity.} By a result in~\cite{Shwartz-David-2014}, our bounds on $\Ndim{\hclass{}}$ in Theorem~\ref{thm:bound-nata} also imply the following bounds on the sample complexity $\sampcom{\hclass{}}$: $(i)$ Lower bound: $c_1 \frac{\sigma + \log(1 / \delta)}{\epsilon}$; $(2)$ Upper bound: $\frac{1}{\epsilon} \cdot \left( c_2 \cdot k\sigma \cdot \log{}(\frac{k\sigma}{\epsilon}) + k \sigma^2 + \log{}(\frac{1}{\delta}) \right)$ for some constants $c_1, c_2 \geq 0$.

\noindent
\rem{} The above upper bound is weaker than our bound in Theorem~\ref{thm:sample}. Notably, the above bound has a dominant term $O(k \sigma^2)$, while our bound in Theorem~\ref{thm:sample} is $O(k\sigma \log{} (k \sigma))$. Further, when $k$ is a constant (a realistic scenario in real-world networks by the Dunbar's number~\cite{dunbar1993coevolution}), the upper bound in Theorem~\ref{thm:sample} is within a factor $O(\log{}(\sigma))$ of the lower bound stated above. 
%%% The following line is not needed.
%This shows a close approximation between our %established upper and lower theoretical limits.

\subsection{The Asymptotic Behavior of Ndim}
\label{sec:asymptotic}
% We further explore the expressive power of the hypothesis class by investigating the limiting distribution of Ndim over all the hypothesis classes.
We have shown that $\Ndim{\hclass{}} \leq k\sigma$ for any $k$-layer system. In this section, we further explore the complexity of the hypothesis class and prove that asymptotically
(i.e., as $n \rightarrow \infty$), for {\em almost all} graphs with $n$ vertices and $k \geq 2$ layers, $\Ndim{\hclass{}}$ of the corresponding hypothesis class is {\em exactly} $k \sigma$,
%i.e., the highest expressiveness possible 
which matches our upper bound
(Theorem~\ref{thm:bound-nata}).

\noindent
\textbf{Approach overview.} Given a multilayer network $\mc{M}$, we first define a special set $\mc{Q}$ of vertex-layer pairs in $\mc{M}$. We then show that for each such set $\mc{Q}$, there is a shatterable set of size $|\mc{Q}|$. Next, using a probabilistic argument, we prove that in the space of all $k$-layer graphs with $n$ vertices, the proportion of graphs that admit such sets $\mc{Q}$ of size $k \sigma$ asymptotically (w.r.t $n$) tends to $1$. Next, consider a graph $\mc{M}$ chosen uniformly at random from the space of all $k$-layer graphs with $n$ vertices. We prove that, asymptotically with probability $1$ (i.e., almost surely), $\mc{M}$ admits such a special set $\mc{Q}$ that contains {\em all the $k \sigma$ vertex-layer pairs}, thus implying the existence of a shatterable set of size $k \sigma$.

\noindent
\textbf{A special vertex-layer set}. For a $k$-layer network $\mc{M}$ and a subset $\Gv{}'$ of vertices, let $\Q{}$ be a set of vertex-layer pairs $(v, i), v \in \mc{V}', i \in [k]$, such that every $(v, i) \in \Q{}$ satisfies the following condition: $N_{\mc{M}}[v, i] \setminus N_{\mc{M}}[v', i'] \neq \emptyset$ for all pairs $(v', i') \in \Q{}, (v', i') \neq (v, i)$, where $N_{\mc{M}}[v, i]$ is the closed neighborhood of $v$ in the $i$th layer of $\mc{M}$. We first establish the correspondence between such a set $\Q{}$ and a shatterable set.

\begin{mybox2}
\begin{lemma}\label{lemma:set-Q}
Given a multilayer network $\mc{M}$ and a subset $\Gv{}'$ of vertices, for each set $\Q{}$, there is a shatterable set of size $|\Q{}|$ for the corresponding hypothesis class over $\mc{M}$, where thresholds of vertices in $\Gv{}'$ are unknown.
\end{lemma}
\end{mybox2}

Our next lemma shows that, asymptotically, almost all graphs $\mc{M}$ with $n$ vertices and $k$ layers have a set $\Q{}$ of size $k\sigma$, where $\Gv{}'$ is a subset of vertices and $\sigma = |\Gv{}'|$.

\begin{mybox2}
\begin{lemma}\label{lemma:all-graphs}
Given $n \geq 1$, $k \geq 2$, and $\Gv{}' \subseteq [n]$, in the space of all $k$-layer graphs with $n$ vertices, the proportion of graphs that admits a set $\Q{}$ of size $k \sigma$ is at least $1 - 4 \cdot (\sigma k)^2 \cdot (\frac{3}{4})^{n}$.
\end{lemma}
\end{mybox2}

\noindent
\textbf{Proof sketch.} Let $\mc{G}_{n, k, 1/2}$ be the space of $k$-layer graphs with $n$ vertices. Let $\mc{M} \sim \mc{G}_{n, k, 1/2}$ be a graph chosen uniformly at random. We first show that the probability of any pair $(v,i)$ violating the condition of $\mc{Q}_{\mc{M}, \mc{V}'}$ is at most $4 \cdot (\frac{3}{4})^{n}$. It follows that w.p. at most $8 \cdot {\binom{\sigma k}{2}} \cdot (\frac{3}{4})^{n} \leq 4 \cdot (\sigma k)^2 \cdot (\frac{3}{4})^{n}$, there exists such an undesirable pair. Consequently, w.p. at least $1 - 4 \cdot (\sigma k)^2 \cdot (\frac{3}{4})^{n}$, $\mc{Q}_{\mc{M}, \mc{V}'}$ contains all the $k \sigma$ pairs. We remark that such probability can be interpreted as the proportion of graphs in the space of all $k$-layer graphs with $n$ vertices. This concludes the proof.
\qed

Collectively, by Lemmas~\ref{lemma:set-Q} and~\ref{lemma:all-graphs}, we conclude that for any $k \geq 2$, asymptotically in $n$, almost all the hypothesis classes of threshold dynamical systems over $k$-layer graphs have Ndim exactly $k\sigma$. 

\begin{mybox2}
\begin{theorem}\label{thm:asymptotica}
Given $k \geq 2$, as $n$ approaches infinity, almost all the hypothesis classes of threshold dynamical systems over $k$-layer graphs have Ndim exactly $\sigma k$.
\end{theorem}
\end{mybox2}

\noindent
\rem{} The term {\em almost all} is standard in probabilistic methods (e.g., see \cite{alon2016probabilistic,erdHos1977chromatic}). Formally, a property holds for almost all graphs if asymptotically (w.r.t $n$) with probability one, a random graph (drawn the space of all $n$-vertex graphs) possesses that property. In our case, such probability is $1 - 4 \cdot (\sigma k)^2 \cdot (\frac{3}{4})^{n} \sim 1-o(1)$, which approaches $1$ quickly due to the exponent of $n$. We also empirically found that for this proportion to be close to $1$, $n$ only needs to be around $1,000$ (see \textbf{Appendix, Section~\ref{sse:sec-7}}).

\section{Experimental Analysis}
We present experimental studies on the relationships between model parameters and the empirical performance of our \pac{} algorithm. Here, we study the performance of the algorithm on different networks~\cite{magnani2013combinatorial, omodei2015characterizing, stark2006biogrid, coleman1957diffusion}, shown in Table~\ref{tab:networks}. In particular, The objective of our experiments is two-fold: $(i)$ We examine the effect of different model parameters on the empirical loss, as these parameters appear in the derived sample complexity bound; $(ii)$ We empirically verify that the loss decreases as more samples are provided.

\begin{table}[H]
\begin{minipage}[c]{0.6\textwidth}
\centering
 \resizebox{0.9\textwidth}{!}{\begin{tabular}{||l c c c c c||} 
 \hline
 \textbf{Dataset} & \text{Type} & $k$ & $n$ & $m$ & \text{Avg. deg.}\\ [0.5ex] 
 \hline 
  \texttt{Aarhus} & Social & $5$ & $61$ & $620$ & $20.33$\\ \hline
  \texttt{CKM-Phy} & Social & $3$ & $246$ & $1,551$ & $12.61$\\ \hline
  \texttt{Multi-Gnp} & Random & $2$ & $500$ & $7,495$ & $15$\\ \hline
  \texttt{PPI} & Biology & $7$ & $900$ & $12,870$ & $28.6$ \\\hline
  \texttt{Twitter} & Social & $2$ & $2000$ & $10,233$ & $10.23$ \\\hline
\end{tabular}}
\end{minipage}
%\label{tab:networks}
\begin{minipage}[c]{0.38\textwidth}
\caption{\textit{List of multilayer networks}. Parameters $k$, $n$, and $m$ are the number of layers, the number of vertices, and the total number of edges in a network, respectively.
}
\label{tab:networks}
\end{minipage}
\end{table}
% \midvs{}

\noindent
\textbf{Training and testing.} For each network, we have a target system $h^*$ where the threshold of each vertex $v \in \mc{V}$ on each layer $i$ is in $[0, \text{deg}_i(v) + 2]$. For each such $h^*$, a training set $\mc{T} = \{(\mc{C}_i, h^*(\mc{C}_i))\}_{i = 1}^q$ is constructed, where each $\mc{C}_i$ is sampled from a distribution $\mc{D}$. We consider \textit{distributions} where the state of each vertex in $\mc{C}_i \in \mc{T}$ is $0$ w.p. $p$ and $1$ w.p. $1 - p$, for a $p \in \{0.1, 0.5, 0.9\}$. Intuitively, a higher $p$ implies a distribution that is more biased towards vertices in state $0$. Our \pac{} algorithm then uses $\mc{T}$ to learn a hypothesis $h \in \hclass{}$ where all the thresholds are inferred. To evaluate the solution quality, we sample $10,000$ configurations from $\mc{D}$, and compute the \textit{empirical loss} $\ell$, which is the fraction of sampled configurations $\mc{C}$'s where $h(\mc{C}) \neq h^*(\mc{C})$. In presenting the results, each data point is the average over $50$ learned hypotheses.

\subsection{Experimental Results}

\begin{figure}[!h]
  \begin{minipage}[c]{0.7\textwidth}
    \includegraphics[width=0.48\textwidth]{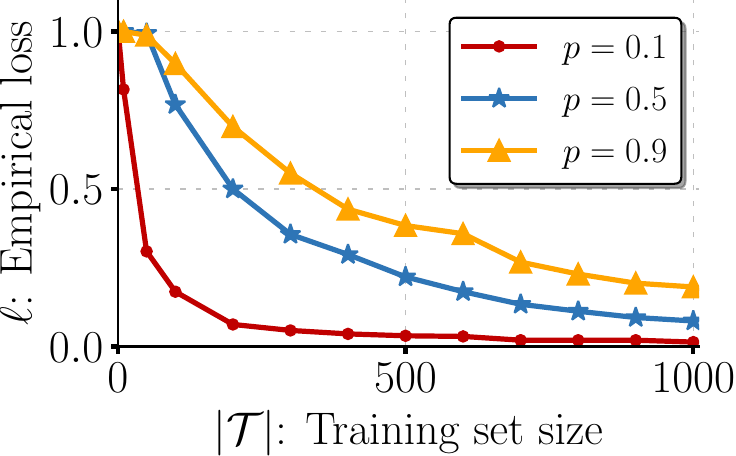}
    \includegraphics[width=0.48\textwidth]{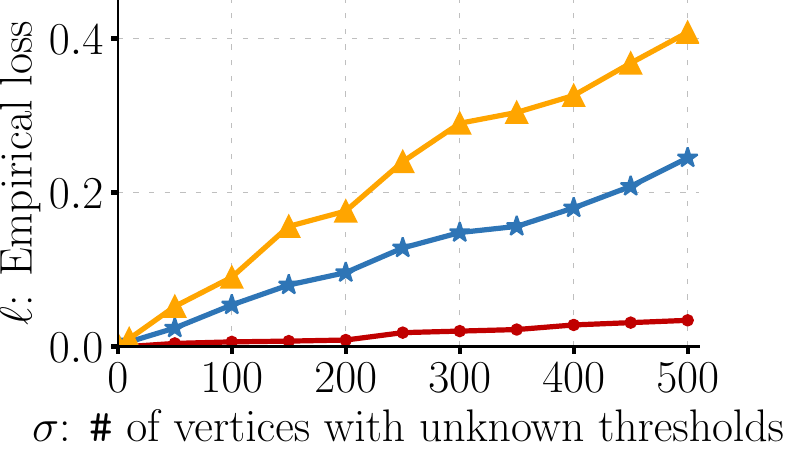}
  \end{minipage}\hfill
  \begin{minipage}[c]{0.3\textwidth}
    \caption{
        (a): $\ell$ vs $|\mc{T}|$ and (b): $\ell$ vs $\sigma$, over different distributions specified by $p$. The underlying network is \texttt{Multi-Gnp} (Table~\ref{tab:networks}). The stdev for all data points is less than $0.09$.
    } \label{fig:gnp_1}
  \end{minipage}
\end{figure}

\textbf{Impacts of model parameters.} We first examine the relationships between the loss $\ell$ and the training set size $|\mc{T}|$, over three distributions specified by different values of $p$. Experimental results using the \texttt{Multi-Gnp} network (Table~\ref{tab:networks}) are in Fig~\ref{fig:gnp_1}(a), where \textit{the interaction functions of all vertices must be learned} (i.e., $\sigma = n$).  

\noindent
\textit{Observations.} In Figure~\ref{fig:gnp_1}(a), the loss $\ell$ decreases as $|\mc{T}|$ increases. Such a relationship is expected since a larger sample usually provides more information about the underlying system. Further, for each value of $|\mc{T}|$, the loss increases as the distribution of samples becomes skewed towards vertices having state $0$. One reason for this behavior is that when states in $\mc{C} \sim \mc{D}$ consist mostly of $0$'s, the scores of vertices under $\mc{C}$ could be far from their true thresholds. Since our algorithm learns based on the scores, it will require more examples to infer an appropriate system. 

Next, we study the relationship between $\ell$ and $\sigma$, under a fixed $|\mc{T}| = 500$ over different distributions. The results for the \texttt{Multi-Gnp} network are shown in Fig~\ref{fig:gnp_1}(b). Specifically, observe that $\ell$ increases with $\sigma$. This is because a larger $\sigma$ leads to an (exponentially) larger hypothesis space. Since the amount of training data (i.e., $|\mc{T}|$) is fixed, a learned hypothesis would incur a higher loss when $\sigma$ is larger. Nevertheless, even though $|\hclass{}|$ is exponential in $\sigma$, the loss $\ell$ of our algorithm grows much more slowly.

\begin{figure}[!h]
  \begin{minipage}[c]{0.7\textwidth}
    \includegraphics[width=0.48\textwidth]{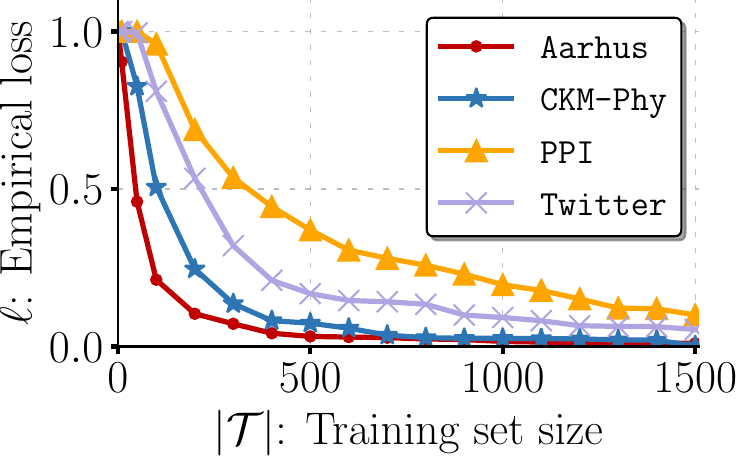}
    \includegraphics[width=0.48\textwidth]{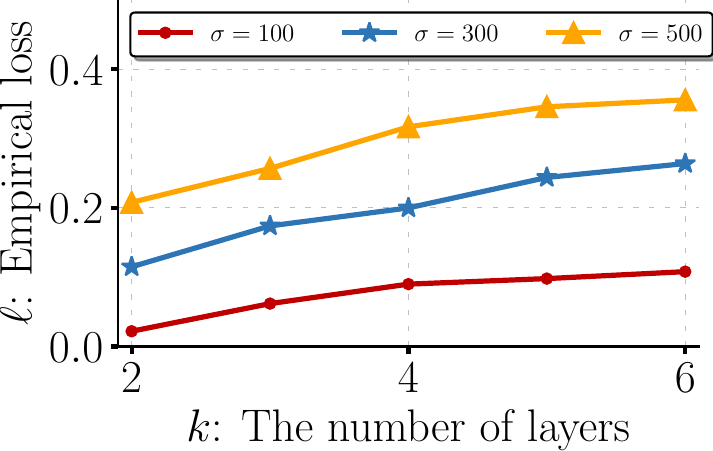}
  \end{minipage}\hfill
  \begin{minipage}[c]{0.3\textwidth}
    \caption{
        (a): $\ell$ vs $|\mc{T}|$, over different real-world networks (Table~\ref{tab:networks}), and (b): $\ell$ vs $k$ over different values of $\sigma$, where the underlying network is \texttt{Gnp}. The stdev for all data points is less than $0.08$.
    } \label{fig:gnp_2}
  \end{minipage}
\end{figure}

\noindent
\textbf{Impact of the number of layers}. We study the effect of the graph structure on $\ell$. We first examine the relationship between $\ell$ and $|\mc{T}|$ using real-world multilayer networks where \textit{the thresholds of all vertices are to be learned}, and $\mc{D}$ is the uniform distribution. The results are in Fig~\ref{fig:gnp_2}(a).

\noindent
\textit{Observations.} From Fig~\ref{fig:gnp_2}(a), we observe a joint effect of $\sigma$ and $k$ on the loss $\ell$. In particular, if the network has more vertices (thus a larger $\sigma = n$), the learned hypothesis $h$ usually has a higher loss $\ell$, as one would expect. Further, even though the \texttt{Twitter} network has more vertices than the \texttt{PPI} network, the latter network has more layers. Since the size of the hypothesis class is exponential w.r.t $k$, for the same $|\mc{T}|$, observe that the $h$ under the \texttt{PPI} network incurs a higher loss. Next, we study the effect of $k$ on the loss $\ell$ using multilayer \texttt{Gnp} networks of size $500$ and average degree (on each layer) of $10$. We increase the number of layers from $2$ to $6$ while \underline{fixing $|\mc{T}|$ at $500$}. The result is shown in Fig~\ref{fig:gnp_2}(b) for three values of $\sigma$. Overall, we observe a positive correlation between $k$ and $\ell$; this is because a larger $k$ leads to a larger hypothesis space.

\section{Future Work}
One direction for future work is to improve our lower bound on the sample complexity for \pac{} learnability. The second direction is to tighten the gap between the lower and upper bounds on the Natarajan dimension for multilayer systems using other techniques. Another promising direction is to consider a noisy setting where labels in the training set (i.e., the successor configurations) may be incorrect with a small probability. Lastly, we note that real-world networks are closer to graphs with special structures, such as multilayer {\em scale-free} or {\em small-world} networks. The multilayer expander graphs and graphs with fixed spectral dimensions are also very interesting due to their theoretical significance. Therefore, we believe that it is important to understand the asymptotic properties of Ndim on such graphs. There are existing works on the asymptotic properties of some other mathematical structures (e.g., cliques~\cite{daly2020asymptotics}, number of triangles~\cite{bollobas2003mathematical}) on scale-free graphs and small-world graphs~\cite{bollobas2003mathematical}. However, we note that the problems studied in these references are very different from our learning problem. Moreover, the networks considered in these references have only a single layer. We believe that obtaining results for Ndim on special classes of multilayer graphs such as multilayer scale-free, small-world, and multilayer expanders poses challenges that are likely to require the development of new theoretical tools. 

% ----------------
%       Bib      -
% ----------------
\bibliographystyle{plain}
\bibliography{bib}

\clearpage

% Appendix
\begin{center}
\fbox{{\Large\textbf{Appendix}}}
\end{center}

\subsection{The Settings of Existing Works}
\label{sse:settings}
Our problem setting follows the line of existing research on learning networked systems. Here, we present the settings of a few illustrative papers on learning networked systems that span multiple authors and domains. These references are also cited in the main paper. 

\begin{table}[H]
\centering
 \resizebox{\textwidth}{!}{\begin{tabular}{||l c c c c ||} 
 \hline
 \textbf{Vertex States} & \textbf{Update Scheme} & \textbf{Time Scale} & \textbf{Interaction Function} & \textbf{Venue}\\ [0.5ex] 
 \hline 
 
   Binary & Synchronous & Discrete & Deterministic & \texttt{AAAI-2022}~\cite{conitzer2022learning}\\ \hline
   
 Binary & Synchronous & Discrete & Threshold & \texttt{ICML-2022}~\cite{rosenkrantz2022efficiently}\\ \hline

 Binary & Synchronous & Discrete & Threshold & \texttt{ICML-2021}~\cite{chen2021network}\\ \hline

 Binary & Synchronous & Discrete & Susceptible-Infected & \texttt{ICML-2021}~\cite{dawkins2021diffusion} \\ \hline

 Binary & Synchronous & Discrete & Independent Cascade & \texttt{ICML-2021}~\cite{wilinski2021prediction} \\ \hline

 Binary & Synchronous & Continuous & Probablistic & \texttt{NeurIPS-2020}~\cite{he2020network}\\ \hline

Binary & Synchronous & Discrete & Threshold & \texttt{NeurIPS-2020}~\cite{li2020online}\\ \hline

 Binary & Synchronous & Discrete & Deterministic & \texttt{ICML-2020}~\cite{conitzer2020learning}\\ \hline

  Binary & Synchronous & Discrete & Threshold & \texttt{ICML-2019}~\cite{adiga2019pac}\\ \hline

    Binary & Synchronous & Discrete & Threshold \& Independent Cascade & \texttt{NeurIPS-2016}~\cite{he2016learning}\\ \hline
    
Binary & Synchronous & Discrete & Threshold \& Independent Cascade & \texttt{NeurIPS-2015}~\cite{narasimhan2015learnability}\\ \hline
\end{tabular}}

\bigskip

\caption{The problem settings of a few illustrative papers on learning networked systems.}

\label{tab:setting}
\end{table}

\subsection{Additional Material for Section~\ref{sec:pac}} \label{sse:sec-3}

\textbf{Duality between \texttt{OR} and \texttt{AND} master functions with respect to learning} 

\noindent
The \texttt{AND} and \texttt{OR} master functions can be treated similarly in our context. For the \texttt{OR} master function, the state of a vertex $v$ is 1 if the interaction function in at least one layer outputs a 1. Similarly, for the \texttt{AND} master function, the state of a vertex $v$ is $0$ if the interaction function on at least one layer outputs a $0$. Due to this duality, all our results for \texttt{OR} master functions carry over to \texttt{AND} master  functions.

\medskip
\noindent
\textbf{Proofs in Section~\ref{sec:sample-complexity}}

Recall that $\tau^{h}_i(v)$ and $\tau^{h^*}_i(v)$ are the thresholds of $v$ on the $i$th layer in a learned system (hypothesis) $h$ and in the true system $h^*$, respectively.  Fix a vertex $v$ and layer $i \in [k]$. 
For a configuration $\mc{C} \sim \mc{D}$, let $B(\mc{C}, v)$ denote the event ``the threshold condition for $v$ is \textbf{not} satisfied in any of the layers under $\mc{C}$ in the true system $h^*$''. 
For an $h \in \mc{H}$ and a configuration $\mc{C}$, let $A(\mc{C}, i, v, h)$ be the event such that $(1)$ the threshold condition of $v$ on the $i$th layer is satisfied under $\mc{C}$ in $h$, and $(2)$ the event $B(\mc{C},v)$ occurs. Formally, $A(\mc{C}, i, v, h)$ is the event  ``$\score{} \geq \tau^{h}_i(v)$ for the given $i$th layer  and $\Gamma_j[\mc{C}, v] < \tau^{h^*}_j(v), \; \forall j \in [k]$''. 

\newpage

\medskip

\begin{mybox2}
\textbf{Lemma~\ref{lemma:sample}.} {\em For a $v \in \mc{V}'$ and an $i \in [k]$, suppose $\tau^{h^*}_i(v) \geq 1$. Let $h \in \hclass{}$ be a hypothesis learned from a training set $\mc{T}$ of size $q \geq 1$. For a given $\alpha \in (0, 1)$,

$(1)$ If all integer $\rho_i(v) \in [0, \tau^{h^*}_i(v))$ satisfy: 
\begin{equation}
    \Pr_{\mc{C} \sim \mc{D}}[\underbrace{\Gamma_j[\mc{C}, v] < \tau^{h^*}_j(v), \; \forall j \in [k]}_\text{Event $B(\mc{C}, v)$} \text{ and } \Gamma_i[\mc{C}, v] \geq \rho_i(v)] < \alpha
\end{equation}
then $\Pr_{\mc{C} \sim \mc{D}}[A(\mc{C}, i, v, h)] < \alpha$. 

(2) If $(1)$ does not hold, that is, there is a $\rho_i(v)$ such that
\begin{equation}\label{eq:rho1}
    \Pr_{\mc{C} \sim \mc{D}}[\underbrace{\Gamma_j[\mc{C}, v] < \tau^{h^*}_j(v), \; \forall j \in [k]}_\text{Event $B(\mc{C}, v)$} \text{ and } \Gamma_i[\mc{C}, v] \geq \rho_i(v)] \geq \alpha
\end{equation}
then the condition $\Pr_{\mc{C} \sim \mc{D}}[A(\mc{C}, i, v, h)] \geq \alpha$ holds with probability \textbf{at most} $(1 - \alpha)^q$ over $\mc{T} \sim \mc{D}^q$.
}
\end{mybox2}

\noindent
\textbf{Proof.} 
We first consider the case where $\Pr{}_{\mc{C} \sim \mc{D}}[B(\mc{C}, v)\text{ and } \Gamma_i[\mc{C}, v] \geq \rho_i(v)] < \alpha$ for all integer $\rho_i(v) \in [0, \tau^{h^*}_i(v))$. This case implies that
\begin{equation}
    \Pr{}_{\mc{C} \sim \mc{D}}[B(\mc{C}, v)\text{ and } \Gamma_i[\mc{C}, v] \geq 0] < \alpha
\end{equation}

Let $h$ be the learned hypothesis by our algorithm. We now argue that $\Pr{}_{\mc{C} \sim \mc{D}}[A(\mc{C}, i, v, h)] < \alpha$. In particular, note that the learned threshold $\tau^{h}_i(v)$ is always in the range $[0, \tau^{h^*}_i(v)]$. If $\tau^{h}_i(v) = \tau^{h^*}_i(v)$, the event $A(\mc{C}, i, v, h)$ does not occur. On the other hand, if $\tau^{h}_i(v) < \tau^{h^*}_i(v)$, then the event $\Gamma_i[\mc{C}, v] \geq \tau^{h}_i(v)$ is contained in the event $\Gamma_i[\mc{C}, v] \geq 0$; thus, 

\begin{align}
    \Pr{}_{\mc{C} \sim \mc{D}}[A(\mc{C}, i, v, h)] &= \Pr{}_{\mc{C} \sim \mc{D}}[B(\mc{C}, v) \text{ and } \score{} \geq \tau^{h}_i(v)]\\
    &\leq \Pr{}_{\mc{C} \sim \mc{D}}[B(\mc{C}, v) \text{ and } \Gamma_i[\mc{C}, v] \geq 0] \\ &< \alpha
\end{align}

\par Now consider the second case as stated in Ineq~\eqref{eq:rho1}. Let $\rho_i(v)$ be the \textbf{maximal} integer in $[0, \tau^{h^*}_i(v))$ such that $\Pr{}_{\mc{C} \sim \mc{D}}[B(\mc{C}, v)\text{ and } \Gamma_i[\mc{C}, v] \geq \rho_i(v)] \geq \alpha$. We now establish the claim:

\begin{claim}\label{claim:key}
    If $\exists \; (\mc{C}, \mc{C}') \in \mc{T}$ s.t. $B(\mc{C}, v)$ occurs and $\Gamma_i[\mc{C}, v] \geq \rho_i(v)$, then the algorithm learns an $h \in \mc{H}$ s.t. $\Pr{}_{\mc{C} \sim \mc{D}}[A(\mc{C}, i, v, h)] < \alpha$.
\end{claim}
Suppose such a pair $(\mc{C}, \mc{C}')$ exists in $\mc{T}$. Let $h$ be the hypothesis returned by our algorithm using $\mc{T}$. Note that since $\Gamma_j[\mc{C}, v] < \tau^{h^*}_j(v), \; \forall j \in [k]$ (i.e., the event $B(\mc{C}, v)$), we must have $\mc{C}'(v) = 0$. By the definition of the PAC algorithm in the main manuscript, it follows that the learned threshold satisfies:
\begin{equation}\label{eq:tauvsrho}
    \tau^{h}_i(v) \geq \rho_i(v) + 1
\end{equation}
Recall that $A(\mc{C}, i, v, h)$ is the event ``$\score{} \geq \tau^{h}_i(v)$ and event $B(\mc{C},v)$ occurs''. If $\rho_i(v) = \tau^{h^*}_i(v) - 1$, then we learned the true threshold (i.e., $\tau^{h}_i(v) = \tau^{h^*}_i(v)$), and thus the event $A(\mc{C}, i, v, h)$ does not occur. On the other hand, if $\rho_i(v) < \tau^{h^*}_i(v) - 1$, then
\begin{align}
    \Pr{}_{\mc{C} \sim \mc{D}}[A(\mc{C}, i, v, h)] &= \Pr{}_{\mc{C} \sim \mc{D}}[B(\mc{C}, v) \text{ and } \score{} \geq \tau^{h}_i(v)]\\
    &\leq \Pr{}_{\mc{C} \sim \mc{D}}[B(\mc{C}, v) \text{ and } \score{} \geq \rho_i(v) + 1] \;\;\;\;\;\;\;\;\\
    &< \alpha
\end{align}
where the last inequality follows from the maximality of $\rho_i(v)$. This establishes the Claim. To complete the argument for the second case, let $\eta = \Pr{}_{\mc{C} \sim \mc{D}}[B(\mc{C}, v) \text{ and } \Gamma_i[\mc{C}, v] \geq \rho_i(v)]$. The probability (over $\mc{T} \sim \mc{D}^q$) that no such configuration $\mc{C}$ exists in $\mc{T}$ is $(1 - \eta)^q \leq (1 - \alpha)^q$, where the inequality follows from Eq~\eqref{eq:rho1}. Therefore, with probability at most $(1 - \alpha)^q$, it holds that $\Pr{}_{\mc{C} \sim \mc{D}}[A(\mc{C}, i, v, h)] \geq \alpha$ for the learned hypothesis $h$. This concludes the proof.\qed

\vspace{10pt}

\begin{mybox2}
\textbf{Theorem~\ref{thm:sample}.} {\em For any $\epsilon, \delta \in (0, 1)$, with a training set of size 
$q = \ceil{{1}/{\epsilon} \cdot \sigma k \cdot \log{}({\sigma k}/{\delta})}$, the proposed algorithm learns a hypothesis $h \in \hclass{}$ such that with probabilty at least $1 - \delta$ (over $\mc{T} \sim \mc{D}^q$),}
\begin{equation}\label{eq:algo-bound}
    \Pr{}_{\mc{C} \sim \mc{D}} [h(\mc{C}) \neq h^*(\mc{C})] < \epsilon
\end{equation}
\end{mybox2}

\noindent
\textbf{Proof.} 
For a vertex $v \in \mc{V}'$, an $h \in \hclass{}$ learned by the proposed algorithm, and a configuration $\mc{C} \sim \mc{D}$, recall that $\Ch{}$ denotes the successor of $\mc{C}$ under the system (hypothesis) $h$; $\Ch{}(v)$ is the state of $v$ in $\Ch{}$. Let $A(\mc{C}, v, h)$ be the bad event where $\Ch{}(v) \neq \Cf{}(v)$, that is, the next state of $v$ predicted by $h$ is wrong. 

\noindent
For any layer $i \in [k]$, by the mechanisms of the PAC algorithm in the main manuscript:
    \begin{equation}
        \tau^{h}_i(v) = \max_{(\mc{C}, \mc{C}') \in \mc{T} : \mc{C}'(v) = 0} \{\Gamma_i[\mc{C}, v]\} + 1.
    \end{equation}
We remark that the learned threshold $\tau^{h}_i(v)$ is {\em at most} the value of the true threshold $\tau^{h^*}_i(v)$. 

\noindent
We first establish the claim:
\begin{claim}\label{claim:wrongstate}
    The event $A(\mc{C}, v, h)$ occurs if and only if $\Ch{}(v) = 1$ and $\Cf{}(v) = 0$.
\end{claim}

\noindent
The necessity is trivially true. To prove sufficiency, we show that the case where $\Ch{}(v) = 0$ and $\Cf{}(v) = 1$ never occurs. Note that if $\Ch{}(v) = 0$, under \texttt{OR} master functions, the threshold condition of $v$ is \textbf{not} satisfied in any layer under $h$. That is, $\Gamma_j[\mc{C}, v] < \tau^{h}_j(v), \; \forall j \in [k]$. Since $\tau^{h}_j(v) \leq \tau^{h^*}_j(v), \forall j \in [k]$ , it follows that the threshold condition of $v$ is also \textbf{not} satisfied 
in any of the layers under $h^*$. Therefore, if $\Ch{}(v) = 0$, then we must have $\Cf{}(v) = 0$. This completes the proof of  Claim~\ref{claim:wrongstate}. 

Based on Claim \ref{claim:wrongstate}, a useful interpretation of the event $A(C, v, h)$ is that the threshold condition of $v$ is satisfied in \textbf{at least one} layer under $\mc{C}$ in $h$, but in the true system $h^*$, the threshold condition of $v$ is \textbf{not} satisfied in any of the layers. 

To arrive at the result in Ineq~\eqref{eq:algo-bound}, we first bound the probability (over $\mc{T} \sim \mc{D}^q$) of learning a bad $h$ where $\Pr{}_{\mc{C} \sim \mc{D}}[A(\mc{C}, i, v, h)] \geq \epsilon / (\sigma k)$. For a $\mc{C} \sim \mc{D}$, and a layer $i \in [k]$, recall that $A(\mc{C}, i, v, h)$ is the event %that the threshold condition of $v$ on the $i$th layer is satisfied under $h$, but, the threshold conditions of $v$ is not satisfied on any of the layers under $h^*$. Formally, $A(\mc{C}, i, v, h)$ is the event such as 
``$\score{} \geq \tau^{h}_i(v)$ and $\Gamma_j[\mc{C}, v] < \tau^{h^*}_j(v), \; \forall j \in [k]$''.
Note that the event $A(\mc{C}, v, h)$ occurs if and only if $A(\mc{C}, i, v, h)$ occurs for at least one layer $i \in [k]$. 
 
For any layer $i \in [k]$, if the true threshold $\tau^{h^*}_i(v) = 0$, the event $A(\mc{C}, i, v, h)$ will never happen as the algorithm always learns the correct threshold, i.e., $\tau^{h}_i(v) = \tau^{h^*}_i(v)$. 
Now suppose $\tau^{h^*}_i(v) \geq 1$. 
We can apply Lemma~\ref{lemma:sample} with $\alpha = \epsilon / (\sigma k)$ 
%(as defined in the Lemma). By Lemma~\ref{lemma:sample}, 
and conclude that with probability \textbf{at most} $(1 - {\epsilon}/({\sigma k}))^q$ over the choices of $q$ examples $\mc{T} \sim \mc{D}^q$, the learned $h$ is ``bad''; that is, $\Pr{}_{\mc{C} \sim \mc{D}}[A(\mc{C}, i, v, h)] \geq {\epsilon}/({\sigma k})$. Overall, when considering all the layers in the network, with probability (over $\mc{T} \sim \mc{D}^q$) at most $k \cdot (1 - \epsilon / (\sigma k))^q$, there exists a layer $i \in [k]$ such that
 \begin{equation}\label{ineq:exist-a-layer}
     \Pr{}_{\mc{C} \sim \mc{D}}[A(\mc{C}, i, v, h)] \geq \frac{\epsilon}{\sigma k}
 \end{equation}

\noindent
Next, we bound the probability (over $\mc{T} \sim \mc{D}^q$) of learning a hypothesis $h \in \mc{H}$ such that 
 $\Pr{}_{\mc{C} \sim \mc{D}}[A(\mc{C}, v, h)] \geq \epsilon / \sigma$, that is, the probability of $h$ predicting wrong next state of $v$ is at least $\epsilon / \sigma$. In particular, note that if $\Pr{}_{\mc{C} \sim \mc{D}}[A(\mc{C}, v, h)] \geq \epsilon / \sigma$, then there must exist a layer $i \in [k]$ such that $\Pr{}_{\mc{C} \sim \mc{D}}[A(\mc{C}, i, v, h)] \geq \epsilon / (\sigma k)$. By our aforementioned argument for Ineq~\eqref{ineq:exist-a-layer}, it follows that with probability (over $\mc{T} \sim \mc{D}^q$) at most $k \cdot (1 - \epsilon / (\sigma k))^q$, 
\begin{equation}
    \Pr{}_{\mc{C} \sim \mc{D}}[A(\mc{C}, v, h)] \geq \frac{\epsilon}{\sigma}
\end{equation}
Lastly, we consider the event where $\Ch{} \neq \Cf{}$ for a configuration $\mc{C} \sim \mc{D}$, that is, the successor of $\mc{C}$ predicted by the learned hypothesis $h$ is wrong. Note that if $\Pr{}_{\mc{C} \sim \mc{D}}[\Ch{} \neq \Cf{}] \geq \epsilon$, then there exists a vertex $v \in \Gv{}'$ such that $\Pr{}_{\mc{C} \sim \mc{D}}[A(\mc{C}, v, h)] \geq {\epsilon} / {\sigma}$, which happens with probability (over $\mc{T} \sim \mc{D}^q$) at most $\sigma k \cdot (1 - \epsilon/(\sigma k))^q$. Setting $q = \ceil{\frac{1}{\epsilon} \cdot \sigma k \cdot \log{}(\frac{\sigma k}{\delta})}$, one can verify that $\sigma k \cdot (1 - \epsilon/(\sigma k))^q \leq \delta$. Overall, when $q = \ceil{\frac{1}{\epsilon} \cdot \sigma k \cdot \log{}(\frac{\sigma k}{\delta})}$, with probability at least $1 - \delta$ over the choices of $q$ examples $\mc{T} \sim \mc{D}^q$, the learned $h \in \hclass{}$ satisfies the condition 
\begin{equation}
    \Pr{}_{\mc{C} \sim \mc{D}}[\Ch{} \neq \Cf{}] < \epsilon.
\end{equation}
This completes the proof. \qed

\vspace{10pt}

\noindent
\textbf{Proofs in Section~\ref{sec:pmac}}
\vspace{10pt}

\begin{mybox2}
\textbf{Theorem~3.4.} {\em For any given $\epsilon, \delta, \beta \in (0, 1)$, with a training set $\mc{T}$ of size $q = \ceil{{1}/{\epsilon} \cdot {1}/{\beta} \cdot k \cdot \log{} (\sigma k/\delta)}$, the  proposed algorithm learns an $h \in \hclass{}$, such that with probability at least $1 - \delta$ over $\mc{T} \sim \mc{D}^q$, $h$ satisfies that $$\Pr{}_{\mc{C} \sim \mc{D}}[W(h(\mc{C}), h^*(\mc{C})) \geq \beta \sigma] ~\leq~ \epsilon$$}
\end{mybox2}

\noindent
\textbf{Proof.}
We follow the analysis in Theorem~\ref{thm:sample}. Let $h \in \mc{H}$ be the hypothesis learned by the algorithm. 
Recall that $A(\mc{C}, v, h)$ is the ``bad'' event where $\Ch{}(v) \neq \Cf{}(v)$ for a vertex $v \in \Gv{}'$ and $\mc{C} \sim \mc{D}$. Using Lemma~\ref{lemma:sample} where we set $\alpha = (\epsilon \beta) / k$, with probability (over $\mc{T} \sim \mc{D}^q$) at most $k \cdot (1 - (\epsilon \beta) / k)^q$, we have $\Pr{}_{\mc{C} \sim \mc{D}}[A(\mc{C}, v, h)] \geq (\epsilon \beta)$ for any vertex $v \in \Gv{}'$. Thus, with probability (over $\mc{T} \sim \mc{D}^q$) at most $\sigma k \cdot (1 - (\epsilon \beta) / k)^q$, there exists a vertex $v \in \mc{V}'$ such that $\Pr{}_{\mc{C} \sim \mc{D}}[A(\mc{C}, v, h)] \geq~ \epsilon \beta$. 

Equivalently, with probability (over $\mc{T} \sim \mc{D}^q$) at least $1 - \sigma k \cdot (1 - (\epsilon \beta) / k)^q$, it holds that $\Pr{}_{\mc{C} \sim \mc{D}}[A(\mc{C}, v, h)] < ({\epsilon \beta})$ for all $v \in \mc{V}'$. Then by the linearity of expectation,
\begin{equation}
    \mathbb{E}_{\mc{C} \sim \mc{D}}[W(h(\mc{C}), h^*(\mc{C}))] < \epsilon \beta \cdot \sigma
\end{equation}

Using Markov Inequality, it follows that with probability (over $\mc{T} \sim \mc{D}^q$) at least $1 - \sigma k \cdot (1 - (\epsilon \beta) / k)^q$,  the learned $h \in \hclass{}$ satisfies
\begin{equation}
    \Pr{}_{\mc{C} \sim \mc{D}}[W(h(\mc{C}), h^*(\mc{C})) \geq \beta \sigma] \leq \epsilon
\end{equation}
Setting $q = \ceil{{1}/{\epsilon} \cdot {1}/{\beta} \cdot k \cdot \log{} (\sigma k/\delta)}$, we have $1 - \sigma k \cdot (1 - (\epsilon \beta) / k)^q ~\geq~ 1 - \delta$. This completes the proof.
\qed

\subsection{Additional Material for Section~\ref{sec:nata}}\label{sse:sec-4}

We first revisit some definitions. 

\medskip

\begin{mybox2}
\begin{definition}[\textbf{Shattering}]
Given a hypothesis class $\mc{H}$, a set $\mc{R} \subseteq \mc{X}$ is \textbf{shattered} by $\mc{H}$ if there exist two functions $g_1, g_2 : \mc{R} \rightarrow \mc{X}$ that satisfy both of the following conditions:
\begin{itemize}[leftmargin=*,noitemsep,topsep=0pt]
    \item Condition 1: For every $\mc{C} \in \mc{R}$,~ $g_1(\mc{C}) \neq g_2(\mc{C})$.
    
    \item Condition 2: For every subset $\mc{R}' \subseteq \mc{R}$, there exists $h \in \mc{H}$ such that $\forall \mc{C} \in \mc{R}', \; h(\mc{C}) = f(\mc{C})$ and $\forall \mc{C} \in \mc{R} \setminus \mc{R}', \; h(\mc{C}) = g(\mc{C})$. 
\end{itemize}
\end{definition}
\end{mybox2}

\begin{figure}[!ht]
  \centering   
  \includegraphics[width=0.8\textwidth]{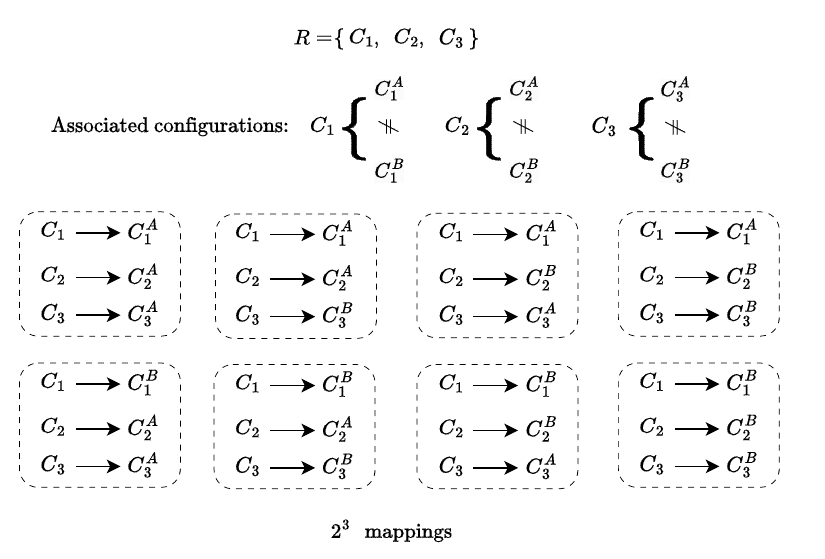}
    \caption{An alternative interpretation of shattering: associated configurations and $2^{|\mc{R}|}$ mappings for a set $\mc{R}$.}
    \label{fig:associated}
\end{figure}

\noindent
\textbf{An alternative interpretation of shattering.} In our context, equivalent definitions of the two conditions are as follows:
\begin{itemize}[leftmargin=*,noitemsep,topsep=0pt]
    \item Condition 1: Each $\mc{C} \in \mc{R}$ is associated with two configurations, denoted by $\mc{C}^A$ and $\mc{C}^B$, where $\mc{C}^A \neq \mc{C}^B$ (i.e., $\mc{C}^A = g_1(\mc{C})$ and  $\mc{C}^B = g_2(\mc{C})$).
    
    \item Condition 2: Consider the $2^{|\mc{R}|}$ possible mappings from $\mc{R}$ to the associated configurations, such that in each mapping $\Phi$, every $\mc{C} \in \mc{R}$ is mapped to one of its associated configuration (i.e., $\Phi(\mc{C}) = \mc{C}^A$ or $\Phi(\mc{C}) = \mc{C}^B$). 
    For each such mapping $\Phi$, there exists a system (hypothesis) $h_{\Phi} \in \hclass{}$ that produces $\Phi$. That is, $h_{\Phi}(\mc{C}) = \Phi(\mc{C})$ for all $\mc{C} \in \mc{R}$. 
\end{itemize}

\medskip

\begin{mybox2}
\begin{definition}[\textbf{Contested Vertices}]
We call a vertex $v$ \textbf{contested} for a $\mc{C} \in \mc{R}$ if $\mc{C}^A(v) \neq \mc{C}^B(v)$.
\end{definition}
\end{mybox2}

\noindent
We use the above definition of shattering in all the proofs. An example of condition 1 and the $2^{|\mc{R}|}$ mappings of condition 2 for a set $\mc{R}$ with three configurations are shown in Fig~\ref{fig:associated}.

\medskip

\begin{mybox2}
\begin{definition}[\textbf{Landmark Vertices}]
    Suppose the underlying network has a single layer. Given a set $\mc{R} \subseteq \mc{X}$, a vertex $v \in \mc{V}'$ is a \textbf{\textit{landmark}} vertex for a configuration $\mc{C} \in \mc{R}$ if $\Gamma[\mc{C}, v] \neq \Gamma[\Hat{\mc{C}}, v]$ for all $\Hat{\mc{C}} \in \mc{R} \setminus \{\mc{C}\}$.
\end{definition}
\end{mybox2}

\noindent
Let $\mc{W}(\mc{R}) \subseteq \mc{V}'$ be the set vertices that are landmarks for at least one configuration in $\mc{R}$.

\medskip

\begin{mybox2}
\begin{definition}[\textbf{Canonical Sets}]
    Suppose the underlying network has a single layer. A set $\mc{R} \subseteq \mc{X}$ is \textbf{\textit{canonical}} w.r.t. $\hclass{}$ if there exists an injective mapping from $\mc{R}$ to $\mc{W}(\mc{R})$ s.t. each $\mc{C} \in \mc{R}$ is mapped to a landmark vertex of $\mc{C}$.
\end{definition}
\end{mybox2}

\noindent
\textbf{Detailed Proofs in Section~\ref{sec:nata}.1}

\medskip
\medskip

\begin{mybox2}
\textbf{Lemma~\ref{lemma:canonical}.} {\em When the underlying network has a single-layer, a set $\mc{R} \subseteq \mc{X}$ can be shattered by $\hclass{}$ \textbf{if and only if} $\mc{R}$ is canonical w.r.t. $\hclass{}$} 
\end{mybox2}

\noindent
\textbf{Proof.} 
$(\Rightarrow)$ Suppose $\hclass$ shatters $\mc{R}$. We want to show that $\mc{R}$ is canonical. For each configuration $\mc{C} \in \mc{R}$, let $\mc{C}^A$ and $\mc{C}^B$ be the two associated configurations, where $\mc{C}^A \neq \mc{C}^B$ (i.e., they disagree on the state of at least one vertex). Consider the $2^{|\mc{R}|}$ possible mappings from $\mc{R}$ to the associated configurations, where in each mapping $\Phi$, each $\mc{C} \in \mc{R}$ is mapped to one of its associated configuration (i.e., $\Phi(\mc{C}) = \mc{C}^A$ or $\Phi(\mc{C}) = \mc{C}^B$). The \textit{second condition} of shattering implies that for each of the mapping $\Phi$ defined above, there exists a system $h_{\Phi} \in \hclass{}$ such that $h_{\Phi}(\mc{C}) = \Phi(\mc{C})$ for all $\mc{C} \in \mc{R}$. 

\par
\begin{figure}[!ht]
  \centering   
  \includegraphics[width=0.25\textwidth]{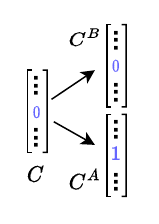}
    \caption{An example of a contested vertex $v$ for a configuration $\mc{C}$. In particular, $\mc{C}^A$ and $\mc{C}^B$ are the two associated configurations of $\mc{C}$. The state of $v$ is highlighted in blue.}
    \label{fig:contested}
\end{figure}

 Recall that a vertex $v$ \textbf{contested} for a configuration $\mc{C} \in \mc{R}$ if the state of $v$ in $\mc{C}^A$ is different from its state in $\mc{C}^B$. An example of a contested vertex is given in Fig~\ref{fig:contested}. Since $\mc{C}^A \neq \mc{C}^B$, each $\mc{C} \in \mc{R}$ has at least one contested vertex. We argue that contexted vertices can only be in $\mc{V}'$.

\begin{claim}\label{claim:onlyVprime}
If $\hclass$ shatters $\mc{R}$, then contested vertices can only be in the set $\mc{V}'$; that is, only vertices with unknown thresholds can be contested.
\end{claim}
For purposes of contradiction, suppose there exists a vertex $v \in \mc{V} \setminus \mc{V}'$ whose threshold is known, and $v$ is contested for a configuration $\mc{C} \in \mc{R}$. The second condition of shattering implies that there exist two systems $h, h' \in \hclass$ such that the state of $v$ is $1$ in $h(\mc{C})$, and is $0$ under $h'(\mc{C})$. However, since the threshold of $v$ is fixed, for the same configuration $\mc{C}$, the state of $v$ is always the same in the successor of $\mc{C}$ regardless of the underlying system in $\hclass{}$. Therefore, such $h, h' \in \hclass$ cannot coexist, which contradicts the fact that $\hclass{}$ shatters $\mc{R}$. 
This establishes the claim.

Our argument of $\mc{R}$ being canonical is developed based on this notion of contested vertices. Overall, we want to show the following two claims: $(i)$ configurations in $\mc{R}$ do not share contested vertices, and $(ii)$ a contested vertex for a $\mc{C} \in \mc{R}$ is also a landmark vertex for $\mc{C}$. Then since each $\mc{C} \in \mc{R}$ has at least one contested vertex, it immediately follows that there exists an injective (i.e., one-to-one) mapping from $\mc{R}$ to $\mc{W}(\mc{R})$ where $\mc{W}(\mc{R}) \subseteq \mc{V}'$ is the set of vertices that are landmarks for at least one configuration in $\mc{R}$. Then by definition, $\mc{R}$ is canonical.

\par We now establish the above two claims. Recall that
for a configuration $\mc{C}$ and a vertex $v$, $\Gamma[\mc{C},v]$ is the \textbf{score} of $v$ in $\mc{C}$, that is, the number of 1's in
the input provided by $\mc{C}$ to the interaction function.

\begin{claim}\label{claim:shatter-1}
    If $\hclass$ shatters $\mc{R}$, then no two configurations in $\mc{R}$
    can have any common contested vertices.
\end{claim}

For purposes of contradiction, suppose $v \in \Gv{}'$ is a contested vertex for at least two configurations in $\mc{R}$; let $\mc{C}_a$ and $\mc{C}_b$ be two such configurations. We now show that $\hclass{}$ cannot shatter $\mc{R}$. Recall that $h(\mc{C}_a)(v)$ is the state of $v$ in the successor $h(\mc{C}_a)$ of $\mc{C}_a$ under a system $h \in \mc{H}$. By the second condition of shattering, there exists a $h \in \hclass{}$ such that $h(\mc{C}_a)(v) \neq h(\mc{C}_b)(v)$ (i.e., $h(\mc{C}_a)(v) = 1$ and $h(\mc{C}_b)(v) = 0$, or $h(\mc{C}_a)(v) = 0$ and $h(\mc{C}_b)(v) = 1$).

If $\Gamma[\mc{C}_a, v] = \Gamma[\mc{C}_b, v]$, then there cannot exist such a system $h \in \hclass{}$ where $h(\mc{C}_a) (v) \neq h(\mc{C}_b)(v)$ since the threshold condition of $v$ cannot be both satisfied and unsatisfied under the same input to the interaction function. This violates the second condition of shattering; thus, $\hclass{}$ fails to shatter $\mc{R}$ under this case. Now suppose $\Gamma[\mc{C}_a, v] < \Gamma[\mc{C}_b, v]$. Then there cannot exist an $h \in \hclass{}$ such that $h(\mc{C}_a)(v) = 1$ but $h(\mc{C}_b)(v) = 0$ since if the threshold condition is satisfied under the smaller score (i.e., $\Gamma[\mc{C}_a, v]$), it must also be satisfied under the larger score (i.e., $\Gamma[\mc{C}_b, v]$). Thus, $\hclass{}$ fails to shatter $\mc{R}$. The argument for the case where $\Gamma[\mc{C}_a, v] > \Gamma[\mc{C}_b, v]$ follows analogously. This concludes the Claim~\ref{claim:shatter-1}. 

%\par Now proceeds to the next claim.
\medskip

\begin{claim}\label{claim:shatter-2}
    If $\hclass$ shatters $\mc{R}$, then a contested vertex for a $\mc{C} \in \mc{R}$ is also a landmark vertex for $\mc{C}$.
\end{claim}

We want to show that if $v \in \mc{V}'$ is contested for $\mc{C} \in \mc{R}$, then $\Gamma[\mc{C}, v] \neq \Gamma[\Hat{\mc{C}}, v]$ for all $\Hat{\mc{C}} \in \mc{R} \setminus \{\mc{C}\}$. Suppose there exists such a  $\Hat{\mc{C}} \in \mc{R}$ where $\Gamma[\mc{C}, v] = \Gamma[\Hat{\mc{C}}, v]$. Claim~\ref{claim:shatter-1} implies that $v$ cannot be contested for $\Hat{\mc{C}}$, that is, the state of $v$ is the same in the two associated configurations of $\Hat{\mc{C}}$; let $s_v$ denote this state value. Given that $v$ is contested for $\mc{C}$, the state of $v$ in one of $\mc{C}$'s associated configurations must be different from $s_v$. Since $\Gamma[\mc{C}_a, v] = \Gamma[\mc{C}_b, v]$, however, there cannot exist an $h \in \hclass$ where $h(\mc{C}_a)(v) \neq s_v$ because the threshold condition of $v$ cannot be both satisfied and unsatisfied under the same input to the interaction function, contradicting the second condition of shattering. This concludes the proof of Claim~\ref{claim:shatter-2}.

Overall, we have shown that configurations in $\mc{R}$ do not share common contested vertices, and that every contested vertex for a configuration is also a landmark vertex. it follows that there exists an injective mapping where each $\mc{C} \in \mc{R}$ is mapped to a landmark vertex of $\mc{C}$. Then by definition, $\mc{R}$ is canonical. 

$(\Leftarrow)$ Suppose that $\mc{R} \subseteq \mc{X}$ is canonical w.r.t $\hclass{}$. To show that $\hclass{}$ shatters $\mc{R}$, we first discuss how the two associated configurations, $\mc{C}^A$ and $\mc{C}^B$, of each $\mc{C} \in \mc{R}$ should be chosen. We then establish that for each of the $2^{|\mc{R}|}$ possible mappings from $\mc{R}$ to the associated configurations (where $\mc{C} \in \mc{R}$ is mapped to one of its associated configurations), there exists a system in $\hclass{}$ that produces this mapping.

\par Given that $\mc{R}$ is canonical, let $\Upsilon: \mc{R} \rightarrow \mc{W}(\mc{R})$ be a corresponding \textit{injective} mapping from $\mc{R}$ to the set $\mc{W}(\mc{R})$ such that each $\mc{C} \in \mc{R}$ is mapped to a landmark vertex of $\mc{C}$. For each $\mc{C} \in \mc{R}$, we construct two associated configuration $\mc{C}^A$ and $\mc{C}^B$ by specifying states of each vertex $v \in \mc{V}$ as follows:

\begin{itemize}[leftmargin=*,noitemsep,topsep=0pt]
    \item \textbf{Case 1}: Suppose $v \in \mc{V} \setminus \mc{V}'$, that is, the threshold of $v$, denoted by $\tau^{h^*}(v)$, is known. Then the state of $v$ in $\mc{C}^A$ and $\mc{C}^B$ is the same, which is determined by $\tau^{h^*}(v)$ and $\Gamma[v, \mc{C}]$. That is, $\mc{C}^A(v) = \mc{C}^B(v) = 1$ if $\Gamma[v, \mc{C}] \geq \tau^{h^*}(v)$, and $\mc{C}^A(v) = \mc{C}^B(v) = 0$ otherwise.
    
    \item \textbf{Case 2}: $v \in \mc{V}'$. 
    \par - \underline{Subcase 2.1}: Suppose $v = \Upsilon(\mc{C})$. Then we set $\mc{C}^A(v) = 0$ and $\mc{C}^B(v) = 1$. That is, $v$ is contested for $\mc{C}$.
    \par - \underline{Subcase 2.2}: Suppose $v \neq \Upsilon(\mc{C})$, and $v = \Upsilon(\Hat{\mc{C}})$ for some other $\Hat{\mc{C}} \in \mc{R}$. Note that the case where $\Gamma[\mc{C}, v] = \Gamma[\Hat{\mc{C}}, v]$ cannot arise since $v$ is a landmark vertex for $\Hat{\mc{C}}$. If $\Gamma[\mc{C}, v] < \Gamma[\Hat{\mc{C}}, v]$, then $\mc{C}^A(v) = \mc{C}^B(v) = 0$. On the other hand, if $\Gamma[\mc{C}, v] > \Gamma[\Hat{\mc{C}}, v]$, then $\mc{C}^A(v) = \mc{C}^B(v) = 1$. 
    \par  - \underline{Subcase 2.3}: Suppose $v \neq \Upsilon(\mc{C})$, and also $v \neq \Upsilon(\Hat{\mc{C}})$ for any other $\Hat{\mc{C}} \in \mc{R}$, then $\mc{C}^A(v) = \mc{C}^B(v) = 1$. 
\end{itemize}
This completes the construction of the two associated configurations $\mc{C}^A$ and $\mc{C}^B$ for each $\mc{C} \in \mc{R}$. We now show that $\hclass{}$ shatters $\mc{R}$ under the defined associations. 

To begin with, observe that $\mc{C}^A \neq \mc{C}^B$ for all $\mc{C} \in \mc{R}$, as the states of $\Upsilon(\mc{C})$ are different in $\mc{C}^A$ and $\mc{C}^B$. Thus, the first condition of shattering is satisfied. 

 Now consider the $2^{|\mc{R}|}$ possible mappings from $\mc{R}$ to the associated configurations, where in each mapping $\Phi$, each $\mc{C} \in \mc{R}$ is mapped to one of its associated configuration (i.e., $\Phi(\mc{C}) = \mc{C}^A$ or $\Phi(\mc{C}) = \mc{C}^B$). To prove the second condition of shattering, we want to show that for each $\Phi$ defined above, there exists a system $h_{\Phi} \in \hclass{}$ such that $h_{\Phi}(\mc{C}) = \Phi(\mc{C})$ for all $\mc{C} \in \mc{R}$. Given a mapping $\Phi$, we characterize $h_{\Phi}$ by presenting how the threshold of each vertex is determined:
\begin{itemize}[leftmargin=*,noitemsep,topsep=0pt]
    \item \textbf{Case 1}: Suppose $v \in \mc{V} \setminus \mc{V}'$, then its threshold is already known.
    \item \textbf{Case 2}: Suppose $v \in \mc{V}'$. 
        \par - \underline{Subcase 2.1}: If $v \neq \Upsilon(\mc{C})$ for any $\mc{C} \in \mc{R}$, then we set $v$'s threshold to be $0$. 
        \par - \underline{Subcase 2.2}: If $v = \Upsilon(\mc{C})$ for a $\mc{C} \in \mc{R}$. Then set $v$'s threshold to be $\Gamma[\mc{C}, v]$ if $\Phi(\mc{C})(v) = 1$. On the other hand, $\Phi(\mc{C})(v) = 0$, then set $v$'s threshold to be $\Gamma[\mc{C}, v] + 1$.
\end{itemize}
This completes the specification of $h_{\Phi}$. One can easily verify that $h_{\Phi}(\mc{C}) = \Phi(\mc{C})$ for all $\mc{C} \in \mc{R}$; that is, the second condition of shattering is satisfied. Overall, we have shown that $\mc{H}$ shatters $\mc{R}$. This concludes the proof. \qed

\vspace{10pt}

\begin{mybox2}
\textbf{Theorem~\ref{thm:exactly-sigma}.}
{\em When the underlying network has a single layer, a shatterable set of size $\sigma$  can be constructed. 
Thus, we have $\Ndim{\hclass{}} = \sigma$.
}
\end{mybox2}

\noindent
\textbf{Proof.} 
We show how a canonical set of size $\sigma$ can be
constructed. Then the theorem follows from the equivalence between a canonical set and a shatterable set.  In particular, given \textit{any} underlying single-layer network $\mc{G}$, we present an algorithm to construct a canonical set $\mc{R} \subset \mc{X}$ that consists of $\sigma$ configurations. 

\par Let $\mc{G}' = \mc{G}[\mc{V}']$ be the subgraph induced on $\mc{V}'$; $\mc{G}'$ could be disconnected. The algorithm involves a depth-first traversal over $\mc{G}'$, starting from any initial vertex. During the traversal, when a vertex $v \in \mc{V}'$ is visited for the first time, a configuration $\mc{C}_v$ is constructed. In particular, our algorithm enforces $v$ to be the landmark vertex to which $\mc{C}_v$ is mapped under the injective mapping defined for a canonical set. 

\begin{figure}[!ht]
  \centering   
  \includegraphics[width=1\textwidth]{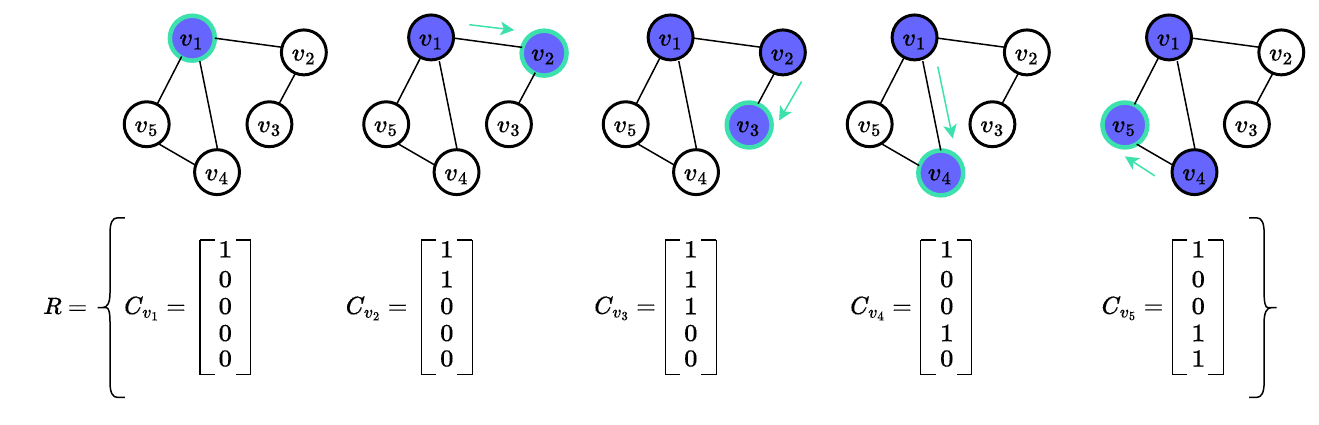}
    \caption{A pictorial example of the algorithm running on a graph of $5$ vertices. Vertices in the stack $\mc{K}$ are  highlighted in blue.}
    \label{fig:dfs}
\end{figure}

We now describe the algorithm. The set $\mc{R}$ is initially empty. Starting from any vertex $v_1 \in \mc{V}'$, we proceed with a depth-first traversal on $\G{}'$, while maintaining a stack $\mc{K} \subseteq \mc{V}'$ of vertices that are currently being visited. Let $v_i$, $i \in [\sigma]$, denote the $i$th vertex that is visited for the first time in the traversal. When $v_i$ is visited for the first time, a configuration $\mc{C}_{v_i}$ is constructed and added to the set $\mc{R}$ where $\mc{C}_{v_i}(v) = 1$ if $v \in \mc{K}$ and $\mc{C}_{v_i}(v) = 0$ otherwise. Note that the states of vertices in $\mc{V} \setminus \mc{V}'$ are always $0$ in $\mc{C}_{v_i}$. The algorithm terminates when all vertices in $\Gv{}'$ are visited (and thus $\mc{K}$ is empty), and returns $\mc{R}$. A pictorial example of the algorithm is given in Fig~\ref{fig:dfs}.

Given the resulting set $\mc{R}$, Since $\mc{G}'$ has $\sigma$ vertices, $|\mc{R}| = \sigma$. We now show that each $v_i \in \Gv{}'$ is a landmark vertex of $\mc{C}_{v_i} \in \mc{R}$, thereby establishing that $\mc{R}$ is canonical. For a $v_i \in \Gv{}'$, recall that the score $v_i$ in a configuration $\mc{C}$, denoted by $\Gamma[\mc{C}, v_i]$, is the number of state-1 vertices in $v$'s closed neighborhood in $\mc{G}$. 

The proof of $v_i$ being a landmark vertex for $\mc{C}_{v_i}$ proceeds in two steps. First, consider the subset $\mc{R}_1 = \{\mc{}{C}_{v_1}, ..., \mc{C}_{v_{i-1}}\} \subset \mc{R}$ of configurations constructed by the algorithm \textit{before} $v_i$ was visited for the first time. (If $v_i = v_1$, then $\mc{R}_1$ is empty.) We argue that the score of $v_i$ in any of the configurations in $\mc{R}_1$ is different from the score of $v_i$ in $\mc{C}_{v_i}$. That is, $\Gamma[\mc{C}, v_i] \neq \Gamma[\mc{C}_{v_i}, v_i]$, $\forall \mc{C} \in \mc{R}_1$. Next, consider the subset $\mc{R}_2 = \{\mc{}{C}_{v_i+1}, ..., \mc{C}_{v_{\sigma}}\} \subset \mc{R}$ of configurations constructed by the algorithm \textit{after} $v_i$ was visited for the first time; if $v_i = v_\sigma$, then $\mc{R}_2$ is empty. Similarly, we argue that the scores of $v_i$ in configurations in $\mc{R}_2$ are different from the score of $v_i$ in $\mc{C}_{v_i}$. We start with the first claim:

\begin{claim}\label{claim:dfs-1}
    For each $i$, $1 \leq i \leq \sigma$,~ $\Gamma[\mc{C}, v_i] \neq \Gamma[\mc{C}_{v_i}, v_i]$, $\forall \mc{C} \in \mc{R}_1$ where $\mc{R}_1 = \{\mc{}{C}_{v_1}, ..., \mc{C}_{v_{i-1}}\} \subset \mc{R}$.
\end{claim}
The claim is trivially true if $v_i = v_1$ since $\mc{R}_1$ is empty. Suppose $i > 1$. Observe that when $v_i$ is visited for the first time, $v_i$ gets added to $\mc{K}$, and thus $\Gamma[\mc{C}_{v_i}, v_i] = \Gamma[\mc{C}_{v_{i-1}}, v_i] + 1$. We now show that before the algorithm visits $v_i$ for the first time, the scores of $v_i$ in the sequence of constructed configurations in $\mc{R}_1$ are non-decreasing. Note that when the algorithm traverses connected components that do \textbf{not} contain $v_i$, the score of $v_i$ is always $0$ in the resulting configurations. Now focus on the connected component containing $v_i$. Recall that in a depth-first traversal, a vertex remains on the stack if it has at least one unvisited neighbor. It follows that all of $v_i$'s neighbors who were visited before $v_i$ will remain on the stack $\mc{K}$ before $v_i$ is visited. Since only vertices on the stack have state-1 in each configuration, the score of $v_i$ is non-decreasing in $(\mc{}{C}_{v_1}, ..., \mc{C}_{v_{i-1}})$, that is $\Gamma[\mc{C}_{v_1}, v_i] \leq ... \leq \Gamma[\mc{C}_{v_{i-1}}, v_i]$. Since $\Gamma[\mc{C}_{v_i}, v_i] = \Gamma[\mc{C}_{v_{i-1}}, v_i] + 1$, it follows that $\Gamma[\mc{C}, v_i] \neq \Gamma[\mc{C}_{v_i}, v_i]$, $\forall \mc{C} \in \mc{R}_1$. This concludes Claim~\ref{claim:dfs-1}.

\noindent
Now, we establish the second claim:

\begin{claim}\label{claim:dfs-2} For each $i$, $1 \leq i \leq \sigma$, $\Gamma[\mc{C}, v_i] \neq \Gamma[\mc{C}_{v_i}, v_i]$, $\forall \mc{C} \in \mc{R}_2$ where $\mc{R}_2 = \{\mc{}{C}_{v_{i+1}}, ..., \mc{C}_{v_{\sigma}}\} \subset \mc{R}$.
\end{claim}

The claim is trivially true if $v_i = v_\sigma$ since $\mc{R}_2$ is then empty. Suppose $i < \sigma$. We show that when a vertex $v_j \in \mc{V}'$, $i < j \leq \sigma$, is visited for the first time (and $\mc{C}_{v_j} \in \mc{R}_2$ is constructed), if $v_i$ is on the stack (i.e., $v_i \in \mc{K}$), then $\Gamma[\mc{C}_{v_j}, v_i] > \Gamma[\mc{C}_{v_i}, v_i]$; if $v_i$ is not on the stack, then $\Gamma[\mc{C}_{v_j}, v_i] < \Gamma[\mc{C}_{v_i}, v_i]$. Let $\mc{N}_{\text{bef}}(v)$ and $\mc{N}_{\text{aft}}(v)$ be the set of neighbors that were visited before and after $v_i$, respectively. Suppose $v_i \in \mc{K}$ when $v_j$ is visited. Note that all vertices in $\mc{N}_{\text{bef}}(v)$ must also be on the stack $\mc{K}$. Further, at least one of $v$'s neighbors in $\mc{N}_{\text{aft}}(v)$ must be on the stack. It follows that $\Gamma[\mc{C}_{v_j}, v_i] \geq \Gamma[\mc{C}_{v_i}, v_i] + 1 > \Gamma[\mc{C}_{v_i}, v_i]$. Now suppose $v_i \notin \mc{K}$ when $v_j$ is visited. This means that no neighbors in $\mc{N}_{\text{aft}}(v)$ are on the stack. It follows that $\Gamma[\mc{C}_{v_j}, v_i] \leq \Gamma[\mc{C}_{v_i}, v_i] - 1 < \Gamma[\mc{C}_{v_i}, v_i]$. Consequently, $\Gamma[\mc{C}, v_i] \neq \Gamma[\mc{C}_{v_i}, v_i]$, $\forall \mc{C} \in \mc{R}_2$. 
This establishes the claim.

\noindent
With Claims~\ref{claim:dfs-1} and \ref{claim:dfs-2}, we have shown that for any $\mc{C}_{v_i} \in \mc{R}$, it holds that
\begin{equation}
    \Gamma[\mc{C}_{v_i}, v_i] \neq \Gamma[\mc{C}, v_i], \; \forall \mc{C} \in \mc{R}, \mc{C} \neq \mc{C}_{v_i}
\end{equation}
That is, $v_i$ is a landmark vertex of $\mc{C}_{v_i}$. This immediately implies the existence of an injective mapping where each $\mc{C}_{v_i} \in \mc{R}$ is mapped to $v_i$, $i \in [\sigma]$. Then by definition, $\mc{R}$ is canonical. Given the equivalence between a canonical set and a shatterable set shown in Lemma~\ref{lemma:canonical}, it follows that $\mc{R}$ is also shatterable by $\hclass{}$. This concludes the proof. \qed

\bigskip

\noindent
\textbf{Detailed Proofs in Section~\ref{sec:nata}.2}

\bigskip

\begin{mybox2}
\textbf{Lemma \ref{lemma:at-most-k-sigma}.}
{\em Suppose the underlying network has $k \geq 2$ layers. Then the size of any shatterable set is at most $k \sigma$.} 
\end{mybox2}

\noindent
\textbf{Proof.} 
We first show that for any shatterable set $\mc{R}$, each vertex $v \in \mc{V}'$ is contested for at most $k$ configurations in $\mc{R}$. Recall that a vertex $v$ is \textbf{contested} for a configuration $\mc{C} \in \mc{R}$ if $\mc{C}^A(v) \neq \mc{C}^B(v)$, where $\mc{C}^A$ and $\mc{C}^B$ are the two associated configurations of $\mc{C}$ defined by shattering (i.e., $\mc{C}^A = g_1(\mc{C})$ and $\mc{C}^B = g_2(\mc{C})$). It is easy to see that Claim~\ref{claim:onlyVprime} in Theorem~\ref{thm:exactly-sigma} carries over to the multilayer case. That is, contested vertices can only be in the set $\mc{V}'$.  

For a $v \in \mc{V}'$, let $\mc{R}_v \subseteq \mc{R}$ be the subset of configurations with $v$ being (one of) their contested vertices. W.l.o.g., suppose $\mc{R}_v \neq \emptyset$. We establish the claim:

\begin{claim}\label{claim:contested-atmostk}
    For each $\mc{C} \in \mc{R}_v$, $\exists \; i \in [k]$ such that $\Gamma_i(\mc{C}, v) > \Gamma_i(\hat{\mc{C}}, v), \; \forall \Hat{\mc{C}} \in \mc{R}_v \setminus \{\mc{C}\}$.
\end{claim}

For contradiction, suppose there exists a $\mc{C} \in \mc{R}_v$ such that for all layers $i \in [k]$, $\Gamma_i(\mc{C}, v) \leq \Gamma_i(\hat{\mc{C}}, v)$ for at least one $\Hat{\mc{C}} \neq \mc{C}$, $\Hat{\mc{C}} \in \mc{R}_v$. We now argue that $\hclass{}$ cannot shatter $\mc{R}_v$ (and thus, cannot shatter $\mc{R}$). In particular, consider a mapping $\Phi$ (among the $2^{|\mc{R}|}$ possible mappings from $\mc{R}$ to the associated configurations) such that $\Phi(\mc{C})(v) = 1$, and $\Phi(\Hat{\mc{C}})(v) = 0$ for all other $\Hat{\mc{C}} \in \mc{R}_v \setminus \{\mc{C}\}$. Suppose there exists a $h_{\Phi} \in \hclass{}$ that is consistent with such a mapping $\Phi$, where $h_{\Phi}(\mc{C})(v) = 1$ and $h_{\Phi}(\Hat{\mc{C}})(v) = 0$ for all $\Hat{\mc{C}} \in \mc{R}_v \setminus \{\mc{C}\}$. Let $\tau^{h_{\Phi}}_i(v)$ denote the threshold of $v$ in the $i$th layer under such an $h_{\Phi}$. Since $h_{\Phi}(\Hat{\mc{C}})(v) = 0$ for all $\Hat{\mc{C}} \neq \mc{C}$, we have

\begin{equation}
    \tau^{h_{\Phi}}_i(v) > \max_{\Hat{\mc{C}} \in \mc{R}_v \setminus \{\mc{C}\}} \Gamma_i(\hat{\mc{C}}, v) \geq \Gamma_i(\mc{C}, v), \forall i \in [k]
\end{equation}

However, the above inequality implies that the threshold condition of $v$ is not satisfied on any of the $k$ layers under $\mc{C}$, thereby contradicting the condition $h_{\Phi}(\mc{C})(v) = 1$. Thus, no such $h_{\Phi} \in \hclass{}$ exists, and $\hclass{}$ does not shatter $\mc{R}_v$. 
This establishes the claim. Overall, Claim~\ref{claim:contested-atmostk} implies that for each $v \in \mc{V}'$, the size of $\mc{R}_v$ is at most $k$. It immediately follows that $|\mc{R}| \leq k  \sigma$ for any shatterable set $\mc{R}$. This concludes the proof.
\qed

\vspace{10pt}

\begin{mybox2}
\textbf{Lemma \ref{lemma:at-least-sigma}.} {\em Suppose $h^*$ is an MSyDS whose underlying network has $k \geq 2$ layers. Let $\Hat{h}^*$ be a single-layer system obtained from $h^*$ by using the network in any layer $i \in [k]$. If a set $\mc{R}$ is shatterable by the hypothesis class of $\Hat{h}^*$, then it is also shatterable by the hypothesis class of $h^*$.}
\end{mybox2}

\noindent
\textbf{Proof.} 
Given the underlying multilayer network $\mc{M} = \{\mc{G}_i\}_{i = 1}^k$ of the true system $h^*$, let $\Hat{h}^*$ be a new system with a single-layer underlying network $\mc{G}_i \in \mc{M}$, for a layer $i \in [k]$. The vertices' thresholds on $\mc{G}_i$ are carried over from $h^*$ to $\Hat{h}^*$. For our learning context, the set $\mc{V}'$ of vertices with unknown thresholds remains the same between $h^*$ and $\Hat{h}^*$. Let $\Hat{\mc{H}}$ be the corresponding hypothesis class of $\Hat{h}^*$,. 

\par Given a set $\mc{R}$ that is shatterable by $\Hat{\mc{H}}$, for each $\mc{C} \in \mc{R}$, let $\Hat{\mc{C}}^A$ and $\Hat{\mc{C}}^B$ be the two associated configurations of $\mc{C}$. 
%under the shatterable condition for $\Hat{\mc{H}}$. 
Further, for each mapping $\Phi$ from $\mc{R}$ to the associated configurations, let $\Hat{h}_{\Phi} \in \Hat{\mc{H}}$ be a system that produces $\Phi$, that is, $\Hat{h}_{\Phi}(\mc{C}) = \Phi(C)$, for all $\mc{C} \in \mc{R}$. 

We show that $\mc{R}$ is also shatterable by $\mc{H}$, the hypothesis class of $h^*$. In particular, for each $\mc{C} \in \mc{R}$, let $\mc{C}^A$ and $\mc{C}^B$ be the two associated configurations under the shatterable condition for $\mc{H}$; we choose $\mc{C}^A = \Hat{\mc{C}}^A$ and $\mc{C}^B = \Hat{\mc{C}}^B$. Now consider any of the $2^{|\mc{R}|}$ mappings from $\mc{R}$ to the associated configurations. We argue that for each of such a mapping $\Phi$, there exists a system $h_{\Phi} \in \mc{H}$ where $h_{\Phi}(\mc{C}) = \Phi(\mc{C})$ for all $\mc{C} \in \mc{R}$. Specifically, in $h_{\Phi}$, the thresholds of each vertex $v \in \mc{V}'$ on each layer $j \in [k]$, denoted by, $\tau^{h_{\Phi}}_j(v)$, are assigned as follows. For each layer $j \in [k]$, if $j = i$ (i.e., the layer for which $\Hat{h}^*$ is defined), then $\tau^{h_{\Phi}}_j(v)$ equals to the threshold of $v$ in $\Hat{h}_{\Phi}$. Otherwise, we set $\tau^{h_{\Phi}}_j(v) = deg_j(v) + 2$, where $\text{deg}_j(v)$ is the degree of $v$ in the $j$th layer. Note that setting $\tau^{h_{\Phi}}_j(v) = \text{deg}_j(v) + 2$ makes $v$'s interaction function on the $j$th layer to be the constant-0 function. One can easily verify that $h_{\Phi}(\mc{C}) = \Phi(\mc{C})$, for all $\mc{C} \in \mc{R}$ and thus $\mc{R}$ is shatterable by $\mc{H}$. 
\qed

\bigskip

\noindent
\textbf{Detailed Proofs in Section~\ref{sec:nata}.3}

\bigskip

\begin{mybox2}
\textbf{Lemma~\ref{lemma:set-Q}.}
{\em 
Given a multilayer network $\mc{M}$ and a subset $\Gv{}'$ of vertices, for each set $\Q{}$, there is a shatterable set of size $|\Q{}|$ for the corresponding hypothesis class over $\mc{M}$ where thresholds of vertices in $\Gv{}'$ are unknown.
}
\end{mybox2}

\noindent
\textbf{Proof.} Given a $k$-layer network $\mc{M}$ with $n$ vertices, let $\Gv{}'$ be any subset of vertices in $\mc{M}$. Let $\mc{H}$ be the threshold dynamical system over $\mc{M}$ where the threshold functions of vertices in $\Gv{}'$ are unknown. For a subset $\Q{}$ of vertex-layer pairs, we present a method to construct a shatterable set $\mc{R}$ of size $|\Q{}|$. In particular, for each $(v, i) \in \mc{Q}_{\mc{M}}$, there is a corresponding configuration $\C{v}{i} \in \mc{R}$, defined as follows: 
\begin{gather*}   
\begin{cases}
  \C{v}{i}(v') = 1 & \text{If } v' \in N[v, i]\\    
  \C{v}{i}(v') = 0 & \text{Otherwise }
\end{cases}
\end{gather*}
where $N[v, i]$ is the closed neighborhood of $v$ on the $i$th layer in $\mc{M}$. 

\par It is easy to see that $|\mc{R}| = |\Q{}|$. We now show that the resulting set $\mc{R}$ is shatterable by $\hclass{}$. Recall that $\Gamma_i[\mc{C}, v]$ is the score (i.e., the number of state-1 vertices in $v$'s closed neighborhood) of $v$ in the $i$th layer under $\mc{C}$. We first observe the following:

\begin{observation}
    $\Gamma_i[\C{v}{i}, v] = deg(v, i) + 1$, for all $(v, i) \in \Q{}$. 
\end{observation}

\begin{observation}
    $\Gamma_{i'}[\C{v}{i}, v'] < deg(v', i') + 1$, for all $(v, i), (v', i') \in \Q{}$, $(v, i) \neq (v', i')$.
\end{observation}

The first observation is held by the construction of $\C{v}{i}$. To see the second observation, recall that $\Q{}$ is defined where $N_{\mc{M}}[v', i'] \setminus N_{\mc{M}}[v, i] \neq \emptyset, \forall (v', i'), (v, i) \in \Q{}, (v', i') \neq (v, i)$. This implies that given a $\C{v}{i} \in \mc{R}$ and a pair $(v' ,i') \in \mc{Q}$, at least one vertex in $N[v', i']$ is in state $0$ under $\C{v}{i}$. The second observation follows immediately.

\noindent
The key conclusion from the above two observations is that:
\begin{align}\label{eq:max-score}
    \Gamma_i[\C{v}{i}, v] > \Gamma_i[\C{v}{i'}, v], \forall i' \neq i, i' \in [k]
\end{align}
This allows us to choose $v$ as the contested vertex for $\C{v}{i}$, $i \in [k]$, under the shattering of $\mc{R}$. For each $(v, i) \in \Q{}$, we now discuss how the two associated configurations of $\C{v}{i}$, denoted by $\C{v}{i}^A$ and $\C{v}{i}^B$, can be chosen to satisfy the shattering conditions. In $\C{v}{i}^A$, the state of $v$ is $1$, where the states of all other vertices are $0$. On the other hand, $\C{v}{i}^B$ is the zero vector. It is clear that $\C{v}{i}^A \neq \C{v}{i}^B$, that is, the first shattering condition is satisfied.

We now show that the second shattering condition also holds. In particular, for each mapping $\Phi$ from $\mc{R}$ to the associated configurations, by choosing the thresholds of vertices, we prove the existence of a system $h_{\Phi}$ that produces the mapping $\Phi$. For each $\C{v}{i} \in \mc{R}$, if $\C{v}{i}^A(v) = 0$, then the threshold of $v$ in the $i$th layer is set to $deg(v, i) + 2$ in $h_{\Phi}$. On the other hand, if $\C{v}{i}^A(v) = 1$, then then the threshold of $v$ in the $i$th layer is set to $deg(v, i) + 1$. By Ineq~\eqref{eq:max-score}, one can easily verify that $h_{\Phi}(\C{v}{i}) = \Phi(\C{v}{i})$, for all $\C{v}{i} \in \mc{R}$. This concludes the proof.
\qed

\bigskip

\begin{mybox2}
\textbf{Lemma~\ref{lemma:all-graphs}.}
{\em  
Given $n \geq 1$, $k \geq 2$, and $\Gv{}' \subseteq [n]$, in the space of all $k$-layer graphs with $n$ vertices, the proportion of graphs that admits a set $\Q{}$ of size $k \sigma$ is at least $1 - 4 \cdot (\sigma k)^2 \cdot (\frac{3}{4})^{n}$.
}
\end{mybox2}

\noindent
\textbf{Proof.}
Recall that $\Q{}$ is a set of vertex-layer pairs $(v, i), v \in \mc{V}', i \in [k]$, such that every $(v, i) \in \Q{}$ satisfies: 
\begin{align}\label{cond:Q}
    N_{\mc{M}}[v, i] \setminus N_{\mc{M}}[v', i'] \neq \emptyset, \forall (v', i') \in \Q{}, (v', i') \neq (v, i)
\end{align}
where $N_{\mc{M}}[v, i]$ is the closed neighborhood of $v$ in the $i$th layer in $\mc{M}$.

We use $\mc{G}_{n, k, 1/2}$ to denote the space of $k$-layer graphs with $n$ vertices. Let $\mc{M}$ be a graph chosen uniformly at random from $\mc{G}_{n, k, 1/2}$, denoted by $\mc{M} \sim \mc{G}_{n, k, 1/2}$. Equivalently, $\mc{M}$ is a random $k$-layer graph with $n$ vertices where {\em each edge in each layer is realized with probability} $p = 1/2$. We will use this equivalent definition in the proof. 

Fix two vertex-layer pairs, $(v, i)$ and $(v', i')$, $v' \in \Gv{}'$, $i \in [k]$, $(v, i) \neq (v', i')$. Let $A \subseteq \mc{V}', \{v, v'\} \subseteq A$, be a subset of vertices that includes $v$ and $v'$. Let $d = A$. Recall that $N[v,i]$ denotes the closed neighborhood of $v$ on the $i$th layer in $\mc{M}$. Suppose $v \neq v'$. The probability (over $\mc{M} \sim \mc{G}_{n, k, 1/2}$) that $N[v,i] = A \text{ and } A \subseteq N[v',i']$ (i.e., condition~\eqref{cond:Q} is violated) is of the form
\begin{align*}
    &\Pr{}_{\mc{M} \sim \mc{G}_{n, k, 1/2}} [N[v,i] = A \text{ and } A \subseteq N[v',i']]\\ &= \underbrace{\frac{1}{2}}_{\text{Edge} (v, v')} \cdot \underbrace{(\frac{1}{2})^{d-2} \cdot (\frac{1}{2})^{n-d}}_{\text{Other neighbors and non-neighbors of }v} \cdot \underbrace{(\frac{1}{2})^{d-2} \cdot \left( \sum_{j=0}^{n-d} {n-d \choose j} (\frac{1}{2})^{j} (\frac{1}{2})^{n - d - j}\right)}_{\text{Other neighbors and non-neighbors of }v'} \\
    &= (\frac{1}{2})^{n + d - 3}
\end{align*}

\noindent
If $v = v'$, then one can verify that $\Pr{}_{\mc{M} \sim \mc{G}_{n, k, 1/2}} [N[v,i] = A \text{ and } A \subseteq N[v',i']] = (1/2)^{n + d - 2}$. 

\noindent
Extending the argument to any such subset $A$, we then have
\begin{align*}
    &\Pr{}_{\mc{M} \sim \mc{G}_{n, k, 1/2}} [N[v,i] \subseteq N[v',i']] \leq \sum_{d = 1}^{n} {n \choose d} (\frac{1}{2})^{n + d - 2}\\
    &<  (\frac{1}{2})^{n - 2} \cdot (\frac{3}{2})^n\\
    &= 4 \cdot (\frac{3}{4})^{n}
\end{align*}

Combining all pairs, the probability (over $\mc{M} \sim \mc{G}_{n, k, 1/2}$) that there exists a $(v, i)$ and $(v', i')$, $v \in \Gv{}', i \in [k]$ such that $N[v,i] \subseteq N[v',i']$ is at most:
\begin{align*}
    8 \cdot {\sigma k \choose 2} \cdot (\frac{3}{4})^{n} \leq 4 \cdot (\sigma k)^2 \cdot (\frac{3}{4})^{n}
\end{align*}

Lastly, with probability at least $1 - 4 \cdot (\sigma k)^2 \cdot (\frac{3}{4})^{n}$, condition~\eqref{cond:Q} holds for all pairs $(v, i), v \in \Gv{}'$, $i \in [k]$, that is, there exists a set $\Q{}$ of size $k \sigma$. This concludes the proof.
\qed

\subsection{Additional Experiments}\label{sse:sec-7}

\textbf{Resources.} All experiments were performed on Intel Xeon(R) Linux machines with 64GB of RAM. Our source code (in \texttt{C++} and \texttt{Python}), documentation, and selected datasets are available in the code appendix.

\noindent
\textbf{Additional Results on the Nararajan Dimension}

Lemma~\ref{lemma:set-Q} in Section~\ref{sec:asymptotic} can be used to estimate the Natarajan dimension of a given graph in the following manner. Two vertex-layer pairs~$(v,i)$ and~$(v',i')$ satisfy the \emph{pairwise non-nested neighborhood} (PNN) property if~$N[v,i]\not\subseteq N[v',i']$ and~$N[v',i']\not\subseteq N[v,i]$. We recall that the Natarajan dimension is lower bounded by the cardinality of any set of vertex-layer pairs satisfying the PNN property. Therefore, our objective here is to find a large set of such pairs. To this end, we construct a graph over vertex-layer pairs, called the PNN graph. We draw an edge between two pairs if they violate the PNN property, i.e., if one of the closed neighborhoods is a subset of another. We apply a greedy vertex coloring algorithm and then choose the largest subset of vertex-layer pairs assigned the same color. By definition of vertex coloring, the chosen vertex-layer pairs form an independent set in the PNN graph, which in turn implies that any vertex-layer pairs in this set satisfy the PNN property. Hence, the cardinality of this set is a lower bound on the Natarajan dimension. 

\begin{figure}[!h]
    \centering    
    \includegraphics[width=.8\textwidth]{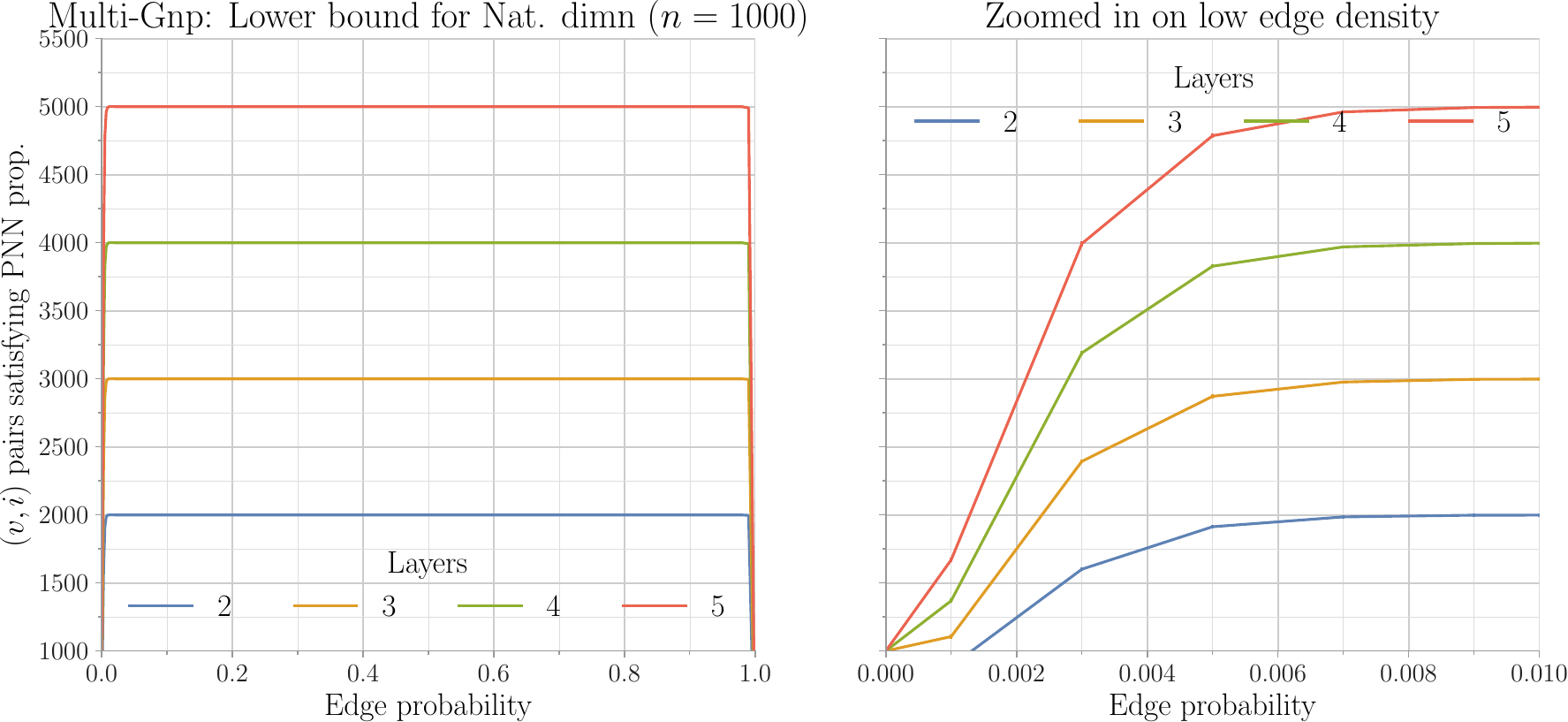}
    \caption{Experimental estimates for Natarajan dimension for Multi-Gnp graphs with a varying number of layers~$k$ and edge probability. Each graph has~1000 vertices. For each value of~$k$ and edge probability,~100 replicates were used. The maximum standard deviation across replicates is less than~50.}
    \label{fig:multignp_pnn}
\end{figure}

The results are shown in Figure~\ref{fig:multignp_pnn}. Recall that the theoretical results show that with very high probability, every vertex-layer pair satisfies the PNN property, and therefore, the Natarajan dimension is~$k\sigma$ w.h.p, even for the case where~$\sigma=n$ for suitably large~$n$. Our experiments suggest that this holds for even smaller graphs, say of size $n=1000$. Secondly, our results indicate that the Natarajan dimension is close to~$k\sigma$ for a large range of edge densities. In the left panel of Figure~\ref{fig:multignp_pnn}, we note that only for very small or very large values of edge
probabilities, the set size reduces. For the extreme cases where the graph is an independent set or is a complete graph in all layers, it can be easily shown that the Natarajan dimension is~$\sigma$.
We observe the same behavior for increasing~$k$.
In the right panel of Figure~\ref{fig:multignp_pnn}, we plot separately for very small edge probabilities to observe the evolution of the lower bound from~$n$ to
$nk$. We note that the standard deviation across replicates is very low as well ($<50$).

\subsection{Additional Remarks and Discussions of the Results}

\noindent
\textbf{Remark on the optimal sample complexity}

\noindent
First, let's recall the following sample complexity bounds from the paper: lower bounded by $\Omega(1/\epsilon \cdot \sigma + 1/\epsilon \cdot \log{}(1/\delta))$
upper bounded by $O(1/\epsilon \cdot \sigma k \cdot \log{}(\sigma k/\delta))$, where $\sigma$ is the number of vertices with unknown interaction functions, and $k$ is the number of layers in the graph. We now elaborate on the intuitions about the optimal sample complexity in both theory and practice.

\begin{itemize}
[leftmargin=*,noitemsep,topsep=0pt]
    \item[-] \textbf{Practice:} The number of layers $k$ in real-world multilayer networks is usually a constant~\cite{dunbar1993coevolution}. In that case, the lower and upper bounds differ only by a factor $O(\log{}(\sigma))$ (Please also see the remark in the main paper for this discussion). This suggests that for the realistic networks that we encounter, the minimum number of samples needed to learn the system will usually be at most a factor $O(\log{}(\sigma))$ smaller than our upper bound above. 

    \item[-] \textbf{Theory:} From a theoretical perspective, without additional assumptions on the sampling distribution, we believe that the factor $k \sigma$ in the sample complexity is inevitable. To see this, note that there are $k \sigma$ unknown interaction functions to be learned. An adversary may choose a distribution (unknown to the learner) such that each data point only reveals useful information to infer at most one such unknown function. Consequently, one needs at least $k \sigma$ samples to infer the full system. Based on this intuition, the optimal sample complexity would be at most a factor $O(\log{}(k \sigma))$ smaller than our upper bound. In general, we note that the issue of determining the optimal sample complexity of multiclass learning is a well-known open problem~\cite{daniely2011multiclass}.
\end{itemize}

\noindent
\textbf{Remark on the difference between learning problems for single-layer vs multi-layer systems}

\noindent
In multilayer dynamics, a state change of a vertex can be caused by the change in the behavior of interaction functions on any of the layers; however, information regarding exactly which of the interaction functions on the layers triggered the change is \textbf{not} contained in the training set. This issue does not occur when the network has only a single layer as there will be \textbf{no} ambiguity.

\begin{figure}[!h]
    \centering    
    \includegraphics[width=\textwidth]{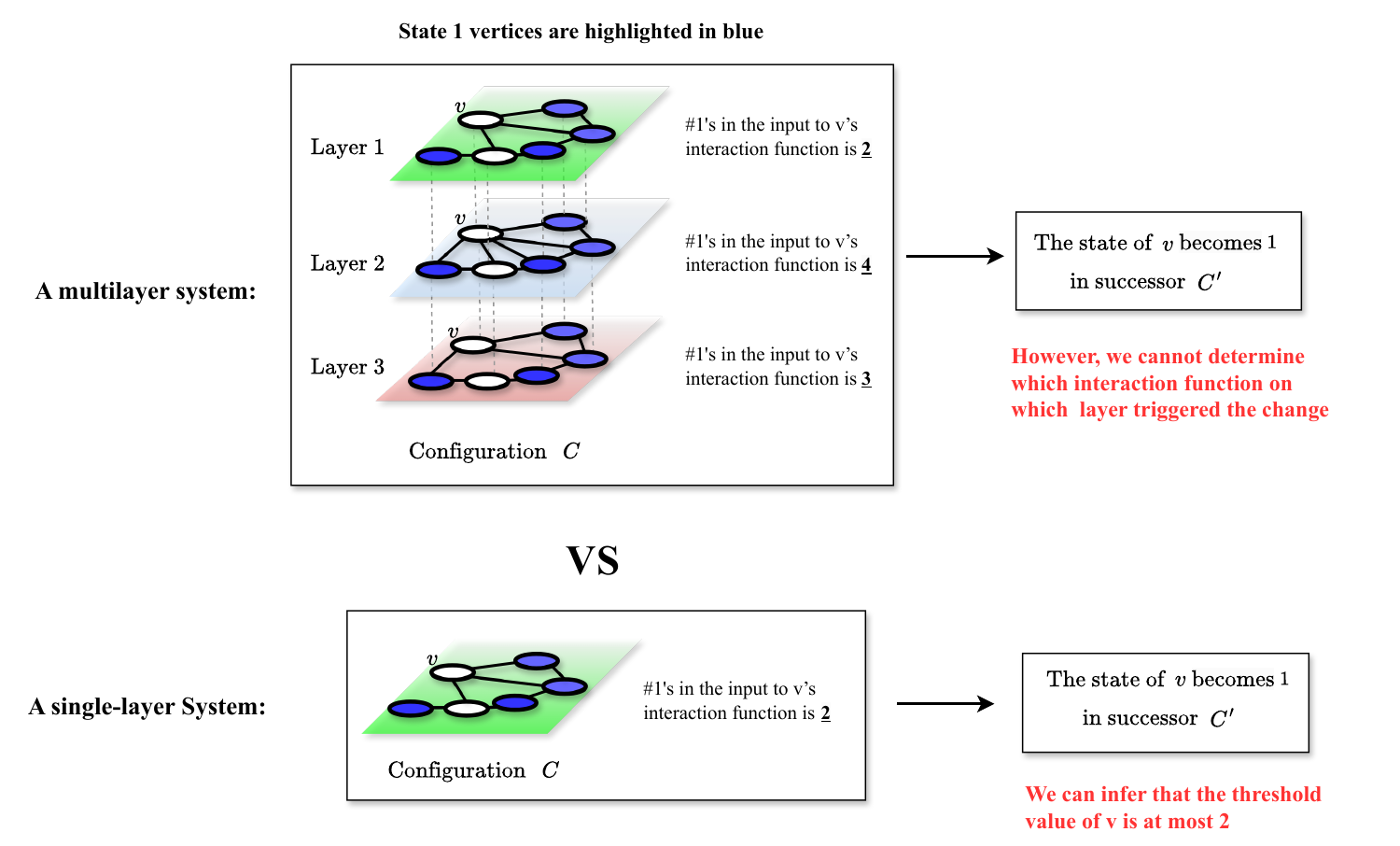}
    \caption{An example of learning a single-layer system vs learing multilayer-system}
    \label{fig:m-vs-s}
\end{figure}

To explain this difference in detail, we provide the following concrete example. The figure of this example is shown in Fig~\ref{fig:m-vs-s}

Consider a single-layer graph $G$ and a multilayer graph $M$ with $3$ layers. In both graphs, we focus on a vertex $v$ whose interaction function(s) are to be learned. Given a data point $(C, C')$ in the training set, please recall that $C'$ is the successor of $C$.

\textbf{Single-layer case}: For a system over the single-layer graph $G$, suppose that under $C$:
\begin{itemize}
    \item The number of 1’s in the input to $v$’s interaction function is $2$
    \item The state of $v$ changes from 0 (under $C$) to 1 (under $C'$).
\end{itemize}
Without ambiguity, this change of state is caused by the input value 2 to $v$’s function; there is only one such function because there is only one layer. So, from this data point $(C, C')$ alone, one can infer that the threshold value of $v$’s function is at most 2. This significantly simplifies the learning process.

\textbf{Multilayer case}: Consider a system over the 3-layer graph $M$. In general, $v$’s interaction functions (which are unknown) can be different on different layers. In this example, suppose that under $C$, the number of 1’s in the input to $v$’s interaction function on each layer is as follows:

\begin{itemize}
    \item 1st layer: the number of 1’s is 2
    \item 2nd layer: the number of 1’s is 4
    \item 3rd layer: the number of 1’s is 3
\end{itemize}

Suppose the state of $v$ changes from 0 (under $C$) to 1 (under $C'$). We remark that all we observe from the data point is the final state of $v$ in $C’$, which is jointly affected by the outputs of the three unknown interaction functions. However, we do not have the actual output values of these (unknown) functions from the training set. So, the difficulty is that we do not know which of the interaction functions on the three layers triggered the change in $v$’s state. Thus, one cannot draw a concrete conclusion on these three unknown interaction functions based only on the entry $(C, C')$. It requires a strategic cross-examination of samples in the entire training set. The issue becomes even more complex as the number of layers $k$ increases.

\end{document}